\pdfoutput=1
\documentclass[10pt,twocolumn]{article}
\usepackage[preprint]{icml2027}

\usepackage[T1]{fontenc}
\usepackage{microtype}
\usepackage{amsmath,amssymb,amsthm,mathtools}
\usepackage{booktabs,multirow,tabularx,array}
\usepackage{graphicx}
\usepackage{xcolor}
\usepackage{algorithm}
\usepackage{algorithmic}
\usepackage{enumitem}
\usepackage{caption}
\usepackage[colorlinks=true,linkcolor=black!70!blue,citecolor=black!70!blue,urlcolor=black!70!blue]{hyperref}
\usepackage[capitalize,noabbrev]{cleveref}
\crefname{atheorem}{Theorem}{Theorems}
\crefname{aproposition}{Proposition}{Propositions}
\crefname{alemma}{Lemma}{Lemmas}
\crefname{acorollary}{Corollary}{Corollaries}
\crefname{adefinition}{Definition}{Definitions}

\setlist{itemsep=1pt,topsep=2pt,parsep=0pt,leftmargin=*}

\theoremstyle{plain}
\newtheorem{theorem}{Theorem}
\newtheorem{proposition}[theorem]{Proposition}
\theoremstyle{remark}

\theoremstyle{plain}
\newtheorem{atheorem}{Theorem}[section]
\newtheorem{aproposition}[atheorem]{Proposition}
\newtheorem{alemma}[atheorem]{Lemma}
\newtheorem{acorollary}[atheorem]{Corollary}
\theoremstyle{definition}
\newtheorem{adefinition}[atheorem]{Definition}

\newcommand{\polar}{\operatorname{polar}}
\newcommand{\F}{\mathrm{F}}
\newcommand{\op}{\mathrm{op}}
\newcommand{\NS}{\mathrm{NS}_5}
\newcommand{\nats}{\,nats}
\newcommand{\aaMax}{0.009}

\newcommand{\absBig}{0.0188}

\newcommand{\absNvM}{0.0150}

\newcommand{\absSNTDvSNT}{0.0052}

\newcommand{\absSNTvS}{0.0158}

\newcommand{\absSNvS}{0.0038}

\newcommand{\absSTvS}{0.0120}
\newcommand{\absSvM}{0.0250}
\newcommand{\actMuona}{$-$0.00054}
\newcommand{\actMuonb}{$+$0.00073}
\newcommand{\actMuonc}{$-$0.00046}

\newcommand{\actStara}{$-$0.00730}
\newcommand{\actStarb}{$-$0.00471}
\newcommand{\actStarc}{$-$0.00183}

\newcommand{\bigLRa}{4.0642}
\newcommand{\bigLRb}{4.0547}
\newcommand{\bigLRc}{4.0647}

\newcommand{\bigSn}{3}
\newcommand{\bigSspread}{0.0008}

\newcommand{\cTopc}{0.98}
\newcommand{\calMax}{5.4}

\newcommand{\ciBig}{[$-$0.0202, $-$0.0175]}

\newcommand{\ciBvS}{[$-$0.0029, $+$0.0051]}

\newcommand{\ciNvS}{[$+$0.0031, $+$0.0170]}

\newcommand{\ciSNTDvSNT}{[$-$0.0066, $-$0.0038]}

\newcommand{\ciSNTvS}{[$-$0.0178, $-$0.0138]}

\newcommand{\ciSNvS}{[$-$0.0062, $-$0.0013]}

\newcommand{\ciSTvS}{[$-$0.0137, $-$0.0102]}
\newcommand{\ciSvM}{[$-$0.0290, $-$0.0210]}

\newcommand{\dBig}{$-$0.0188}

\newcommand{\dBvS}{$+$0.0011}

\newcommand{\dNvS}{$+$0.0100}

\newcommand{\dSNTDvS}{$-$0.0210}
\newcommand{\dSNTDvSNT}{$-$0.0052}

\newcommand{\dSNTvS}{$-$0.0158}
\newcommand{\dSNTvSN}{$-$0.0120}
\newcommand{\dSNTvST}{$-$0.0038}

\newcommand{\dSNvS}{$-$0.0038}

\newcommand{\dSTvS}{$-$0.0120}
\newcommand{\dSvM}{$-$0.0250}
\newcommand{\fixGaina}{$-$0.00589}
\newcommand{\fixGainb}{$-$0.00488}
\newcommand{\fixGainc}{$-$0.00009}
\newcommand{\fixSharea}{87}
\newcommand{\fixShareb}{92}

\newcommand{\gammaa}{0.52}
\newcommand{\gammab}{0.49}

\newcommand{\impA}{$-$0.8}
\newcommand{\impB}{$-$0.8}
\newcommand{\impC}{$-$1.4}
\newcommand{\levAttn}{0.02}
\newcommand{\levCon}{0.18}
\newcommand{\levConSq}{0.02}
\newcommand{\levExp}{0.18}
\newcommand{\levExpSq}{0.03}
\newcommand{\levMomQKV}{0.93}
\newcommand{\levQKV}{0.43}
\newcommand{\levQKVend}{0.47}
\newcommand{\levQKVhi}{0.73}
\newcommand{\levQKVlo}{0.14}
\newcommand{\levQKVmax}{1.84}
\newcommand{\levQKVsq}{0.22}

\newcommand{\minEigAll}{4.6\times10^{-8}}
\newcommand{\minM}{23.3}

\newcommand{\minS}{24.9}
\newcommand{\minSN}{25.1}
\newcommand{\minSNT}{27.8}

\newcommand{\minST}{26.4}

\newcommand{\msGainFixa}{$-$4.5}
\newcommand{\msGainFixb}{$-$3.4}

\newcommand{\msGainTopa}{$-$4.4}
\newcommand{\msGainTopb}{$-$3.2}

\newcommand{\msGapDiffa}{1.4}
\newcommand{\msGapDiffb}{1.5}

\newcommand{\msRealizeda}{77}
\newcommand{\msRealizedb}{69}

\newcommand{\msSlopea}{$+$0.090}
\newcommand{\msSlopeb}{$+$0.060}

\newcommand{\nBigWords}{three seeds}

\newcommand{\nTaila}{7.1}

\newcommand{\nTopa}{1.5}

\newcommand{\nsharea}{5}

\newcommand{\nsharec}{65}
\newcommand{\preMuona}{$-$0.00079}
\newcommand{\preMuonb}{$+$0.00026}
\newcommand{\preMuonc}{$-$0.00028}
\newcommand{\preStara}{$-$0.00753}
\newcommand{\preStarb}{$-$0.00506}
\newcommand{\preStarc}{$-$0.00165}

\newcommand{\psBig}{$-$0.0188, $-$0.0183, $-$0.0194}

\newcommand{\psSTvS}{$-$0.0113, $-$0.0127, $-$0.0119}

\newcommand{\scrBlocks}{$-$0.0219}

\newcommand{\scrBlocksDeploy}{$-$0.0200}

\newcommand{\scrBlocksExact}{$-$0.0332}

\newcommand{\scrFullEarly}{$-$0.0007}
\newcommand{\scrFullHalf}{$+$0.0009}

\newcommand{\scrHeads}{$-$0.0172}

\newcommand{\scrLRc}{$+$0.0034}

\newcommand{\scrLRd}{$+$0.0332}

\newcommand{\scrMLPBlocks}{$+$0.0039}

\newcommand{\scrMuonRange}{0.0032}
\newcommand{\scrNor}{$-$0.0189}

\newcommand{\scrPolish}{$-$0.0020}

\newcommand{\scrPolyBad}{$-$0.0240}

\newcommand{\scrPolyGood}{$-$0.0322}

\newcommand{\scrRowInst}{$-$0.0192}

\newcommand{\scrSchFour}{$+$0.0177}

\newcommand{\scrSchOneCI}{[$-$0.0100, $-$0.0014]}
\newcommand{\scrSchOneMean}{$-$0.0057}

\newcommand{\scrSplit}{$-$0.0255}

\newcommand{\scrToner}{$-$0.0068}

\newcommand{\scrWhNat}{$+$0.0524}

\newcommand{\sharea}{17}
\newcommand{\shareb}{12}
\newcommand{\sharec}{97}

\newcommand{\slopeS}{0.260}
\newcommand{\tanga}{1.00}
\newcommand{\tangb}{1.00}
\newcommand{\tangc}{0.79}
\newcommand{\tiltPeak}{$-$0.050}

\newcommand{\timeBig}{$+$12}

\newcommand{\timeSNTDvS}{$+$26}
\newcommand{\timeSNTvS}{$+$12}

\newcommand{\timeSvM}{$+$7}

\newcommand{\tokSNTvS}{4.3}

\newcommand{\tokSvM}{6.8}

\icmltitlerunning{Equal Steps, Unequal Curvature: Where Muon's Update Goes Wrong}
\icmlauthorblock{%
  {\bfseries Biswa Sengupta\qquad Jinhua Wang\qquad L\'eo Maxime Brunswic}\\[2pt]
  LLM Suite Team, JPMorgan Chase \& Co.\\[1pt]
  \texttt{\{biswa.sengupta, jinhua.wang, leo.brunswic\}@jpmorgan.com}}
\icmlcopyright{\copyright\ 2026 JP Morgan Chase \& Co. All rights reserved}

\begin{document}

\twocolumn[
\icmltitle{Equal Steps, Unequal Curvature:\\ Where Muon's Update Goes Wrong}
\begin{icmlauthorlist}
\icmlauthor{Biswa Sengupta}{jpm}
\icmlauthor{Jinhua Wang}{jpm}
\icmlauthor{L\'eo Maxime Brunswic}{jpm}
\end{icmlauthorlist}
\icmlaffiliation{jpm}{LLM Suite Team, JPMorgan Chase \& Co.}
\icmlcorrespondingauthor{Biswa Sengupta}{biswa.sengupta@jpmorgan.com}
\icmlkeywords{Muon, optimization, orthogonalized updates, curvature, language models, Lean}
\ificmlaccepted
\begin{icmldisclaimer}
This paper was prepared for informational purposes by the LLM Suite group of JP Morgan Chase and its affiliates (`JPMC') and is not a product of the Research Department of JP Morgan. JP Morgan makes no representation, warranty or undertaking whatsoever and disclaims all liability for the completeness, accuracy or reliability of the information contained herein. This document is not intended as investment research or investment advice, or a recommendation, offer or solicitation for the purchase or sale of any security, financial instrument, financial product or service, or to be used in any way for evaluating the merits of participating in any transaction, and shall not constitute a solicitation under any jurisdiction or to any person, if such solicitation under such jurisdiction or to such person would be unlawful.
\end{icmldisclaimer}
\fi
\vskip 0.05in
]
\printAffiliationsAndNotice{}

\begin{abstract}
Muon updates each hidden weight matrix along the polar factor of its momentum, the steepest-descent direction under the spectral norm. That choice fixes two allocations without consulting the loss: every singular direction of the momentum receives the same step, and in a non-square matrix each neuron's step is set by its leverage score. Whether these allocations are good is a question about curvature, which steepest-descent arguments do not answer. We measure it. Using held-out gradients and Hessian--vector products, we decompose Muon's step by singular-value band and by neuron group and compute each group's second-order optimal multiplier. In a 124M-parameter decoder trained on FineWeb-Edu, the multipliers rise monotonically from the top singular directions to the tail: Muon over-steps the high-curvature top of its spectrum. We prove in Lean when such a diagnostic licenses an intervention at Muon's own update norm. Applied to the measured curvature, the criterion predicts, and direct measurements of the held-out loss confirm, that moving step length off the top improves the one-step loss severalfold mid-training and at the start of the cooldown, but not at its end. We then test interventions under per-step norm matching, tuned baselines and paired seeds. Orthogonalizing Q, K and V separately rather than as one fused matrix lowers the loss by \absSvM\nats{} and accounts for all of the gain of block-wise variants. A matmul-only, rotation-equivariant spectral tilt $\sigma^{-1/4}$ applied in the cooldown lowers the loss of this stronger baseline by a further \absSTvS\nats{}, additively with NorMuon (\absSNTvS\nats{} together, three seeds; the baseline needs about \tokSNTvS\% more tokens to match), and by \absBig\nats{} at 355M parameters (\nBigWords). Energy-matched Schatten-1 weight decay adds a further \absSNTDvSNT\nats. Geometrically principled changes that do not act on the measured miscalibration, namely exact polar factors, steepest-descent input whitening and top-heavy decay, do not help.
\end{abstract}

\section{Introduction}\label{sec:intro}

Muon~\citep{muon} has become the reference optimizer for the hidden matrices of transformers at small and medium scale~\citep{moonlight,essentialai,moddednanogpt}. Its theory is clean. For a weight matrix with momentum $N=U\Sigma V^\top$, the update direction $\polar(N)=UV^\top$ is the steepest-descent direction under the spectral norm~\citep{bernstein,modularnorm,scion}, and Muon with decoupled weight decay solves a spectral-norm-constrained problem~\citep{chenlili}.

That theory picks the \emph{norm} in which the step is steepest. It is silent on whether the step's \emph{distribution} is right, and $UV^\top$ fixes two distributions without asking. (i)~\emph{Across singular directions}, every direction $u_iv_i^\top$ receives a step of the same length, whatever its gradient signal $\sigma_i$ and whatever the loss curvature along it. (ii)~\emph{Across neurons}, a tall polar factor has row norms equal to the square roots of the leverage scores of $N$'s column space. Each output neuron therefore moves by an amount set by the momentum's geometry and by how matrices happen to be fused in the implementation.

Recent variants change both distributions. NorMuon, Muon+ and Aurora rebalance neurons~\citep{normuon,muonplus,aurora}. DynMuon, Freon/Kaon and Spectrally Targeted Muon reshape the spectrum~\citep{dynmuon,freon,stmuon}. MuonBP and HiMuon orthogonalize blocks~\citep{muonbp,himuon}. Some work also measures why. SAMuon probes held-out loss-optimal step sizes along the momentum's singular directions and builds an optimizer from that profile~\citep{samuon}. BulkBoost proves when a norm-matched two-band reallocation helps to first order~\citep{bulkboost}. Curvature analyses explain Muon's advantage through its actual update~\citep{wangcurv,freon,quadlens}. These measurements treat directions or bands one at a time or compare Muon with Adam, and gains against loosely tuned baselines often shrink against well-tuned ones~\citep{fantastic,deconstructing}.

\textbf{This paper measures the joint picture.} We treat the flat spectrum and the leverage-weighted neuron allocation as hypotheses and test them against the loss's local geometry. We split Muon's next step into groups of singular directions or of neurons. On held-out data we measure each group's gradient alignment $g_b$ and the full curvature matrix $H$ between groups, including the cross-group terms that isolated probes omit. The ratio $g_b/H_{bb}$ is the step multiplier that would be optimal for group $b$ on its own to second order, and Muon corresponds to multiplier 1 everywhere. A one-step greedy analysis is known to misjudge the overall step size~\citep{shorthorizon}. We therefore use an exact identity, proved in Lean, that splits Muon's second-order gap into a \emph{scale} part (the overall step size) and a \emph{shape} part (the relative weighting across groups). Our experiments compare steps of equal norm, so we also prove when a same-norm reallocation can beat Muon's step, and we evaluate that criterion on the measured curvature.

\textbf{What the anatomy shows} (\cref{sec:anatomy}). The optimal multiplier rises monotonically from the top singular directions to the tail. On ranks 513--768 it is 38--190 times its value on ranks 1--8. Muon thus over-steps the high-curvature top of its spectrum relative to the rest. At Muon's own update norm, which is the comparison our experiments make, reallocating step length off the top would change the one-step loss by \fixGaina{} mid-training and \fixGainb{} at the start of the cooldown. Muon's own step changes it by \preMuona{} and \preMuonb. By the end of the cooldown the possible gain is only \fixGainc. Measuring the actual held-out loss confirms this: at matched norm, halving only the top eight singular directions improves Muon's one-step loss by \msGainTopa{} and \msGainTopb{} ($\times10^{-3}$), and not at the end. Per-neuron step sizes vary most in the fused $2304\times768$ QKV matrix (coefficient of variation \levQKV, against \levExp{} in the MLP).

\textbf{What acting on it buys} (\cref{sec:experiments}). We test interventions under strict controls. Each direction-changing arm is norm-matched to Muon at every step, baselines use tuned (interior-optimum) learning rates, and seeds are paired. Three findings stand out.
\begin{enumerate}
\item \emph{The neuron allocation that matters is per projection.} Orthogonalizing Q, K and V separately, a known practice~\citep{moddednanogpt}, changes the final loss by \dSvM\nats{} \ciSvM{} at 100.7M tokens. It accounts for \emph{all} of the gain of square-block variants (residual \dBvS{} \ciBvS). Finer rebalancing, per head or per row, does worse. The fused-QKV baseline would need about \tokSvM\% more tokens to match, which confounds comparisons: NorMuon's gain shrinks from \absNvM{} to \absSNvS{} against the split baseline.
\item \emph{Tilting the spectrum away from its top helps.} A cooldown tilt toward small singular values lowers the stronger baseline's loss by a further \absSTvS\nats{} (\dSTvS{} \ciSTvS), better in every seed. We implement it with a Chebyshev filter evaluated by Clenshaw's recurrence, using matrix products only, and prove that it keeps Muon's rotational equivariance.
\item \emph{The two corrections add.} NorMuon $+$ tilt gives \dSNTvS{} \ciSNTvS, against \dSNvS{} and \dSTvS{} separately. At 355M parameters the combination gains \absBig\nats{} \ciBig{} (\nBigWords). Energy-matched Schatten-1 weight decay adds a further \dSNTDvSNT{} \ciSNTDvSNT{} at 124M.
\end{enumerate}
The negatives are equally informative. Input whitening is provably $\mathrm{GL}(n)$-equivariant and conditioning-free on least squares, yet it hurts (up to $+0.06$\nats). More accurate polar factors do nothing, and top-heavy decay hurts; none acts on the miscalibration we measured. The diagnostic and the first screens ran concurrently, so the anatomy's implications are consistency checks, not pre-registered predictions; the factorial ablations and the 355M runs came later.

\textbf{Contributions.}
\textbf{(1)}~A held-out, curvature-resolved diagnostic for a matrix optimizer's step, with an exact scale/shape decomposition and a proved criterion for when a same-norm reallocation can improve the step.
\textbf{(2)}~An anatomy of Muon: monotone over-stepping of the top singular directions, which a same-norm reallocation can exploit until late in the cooldown, and a neuron imbalance concentrated in fused matrices.
\textbf{(3)}~A controlled attribution, including a $2{\times}2$ factorial with three seeds and an A/A noise check.
\textbf{(4)}~Machine-checked statements of the geometry used (\cref{app:lean}), and an account of what they did and did not predict (\cref{sec:lean}).
Every optimizer ingredient we evaluate has a precedent, and so does probing Muon's spectrum against held-out loss (\cref{sec:related}). What is new is the joint-curvature diagnostic with its fixed-norm criterion, the neuron-level and QKV attribution, and the controls that test them.

\section{Two allocations hidden in Muon's step}\label{sec:background}

For a hidden matrix $W\in\mathbb R^{m\times n}$ with gradient $G_t$, the Muon variant we study forms the sum-form Nesterov input and steps along a quintic Newton--Schulz map:
\begin{equation}\label{eq:muon}
\begin{aligned}
B_t&=\beta B_{t-1}+G_t,\qquad N_t=G_t+\beta B_t,\\
W_{t+1}&=(1-\eta_t\lambda)W_t-\eta_t\,s\,\NS(N_t),
\end{aligned}
\end{equation}
with shape factor $s=\sqrt{\max(1,m/n)}$. $\NS$ normalizes $N$ by its Frobenius norm and applies five iterations of $X\mapsto aX+b\,XX^\top X+c\,(XX^\top)^2X$ with $(a,b,c)=(3.4445,-4.775,2.0315)$~\citep{muon}. On singular values it acts as a scalar map $\phi_5$ that is close to~1 on most of the spectrum but well below 1 on near-null directions. Hence $\NS(N)\approx\polar(N)=UV^\top=\arg\max_{\|D\|_\op\le1}\langle N,D\rangle$.

\paragraph{Allocation across singular directions.} Write any spectral filter as $D_f=U f(\Sigma)V^\top$. SGD is $f(\sigma)=\sigma$, Muon is (approximately) $f\equiv 1$, and power-law tilts are $f(\sigma)=(\sigma/\sigma_{\rm rms})^p$~\citep{dynmuon}. Muon's choice is optimal for the worst case over a spectral-norm ball. It is not obviously optimal for a loss whose curvature differs across singular directions.

\paragraph{Allocation across neurons.} Every theorem and proposition in this paper is proved in Lean~4; \cref{app:lean} states each in full. Row $i$ of $W$ is the incoming weight vector of output neuron~$i$. Muon moves it by $\|(UV^\top)_{i,:}\|$.
\begin{proposition}[Polar row norms are leverage scores; \cref{prop:a-rows}]\label{prop:leverage}
Let $O\in\mathbb R^{m\times n}$ with $O^\top O=I_n$ ($m\ge n$). Then $\sum_i\|O_{i,:}\|^2=n$, but the individual $\|O_{i,:}\|^2$, the leverage scores of $\operatorname{range}(O)$~\citep{leverage}, can be arbitrarily uneven: one row can have norm 1 and another 0. If $m=n$ and $OO^\top=I$, every row has unit norm.
\end{proposition}
So Muon's per-neuron step depends on the \emph{shape} of the matrix it is applied to. Storing Q, K and V as one fused $3d\times d$ matrix (the GPT-2 layout~\citep{gpt2}) makes each projection's step depend on the momenta of the other two. The block-wise alternative is the exact steepest step of a natural norm:
\begin{proposition}[Block-wise polar steps; \cref{app:lean-blocks}]\label{prop:blocks}
Split $W=[W_1;\dots;W_k]$ into $k$ blocks and let $\|D\|_{\rm blk}=\max_b\|D_b\|_\op$. Then $[\polar(N_1);\dots;\polar(N_k)]$ maximizes $\langle N,D\rangle$ over $\|D\|_{\rm blk}\le1$, and the joint polar factor can be strictly worse (for $N_1=N_2=(1,0)$ it attains $\sqrt2$ against~$2$). For a tall $O$ split into $k$ exactly orthogonal $n\times n$ row blocks, Muon's shape factor $s=\sqrt k$ gives the joint step the same Frobenius energy as the block-wise step.
\end{proposition}
Both propositions concern exact polar factors. $\NS$ is only approximately orthogonal, so in practice block-wise steps shrink the row spread but do not remove it. The block-wise steepest step for this max-norm is MuonBP's BlockMuon~\citep{muonbp}. The leverage-score reading of tall polar factors is also given by \citet{aurora}. We use both diagnostically, to locate where Muon's per-neuron steps are uneven, not to show that the unevenness hurts.

\section{A curvature-resolved anatomy of the step}\label{sec:method}

\paragraph{Grouped multipliers.} We partition Muon's next step into groups $b=1,\dots,B$ in one of two ways. \emph{Bands:} singular-value ranks of the momentum, with $D_{\ell,b}=\eta\,s_\ell\,U_{I_b}V_{I_b}^\top$ for every matrix $\ell$. This is the exact-polar idealization of Muon's step; $\NS$ itself gives near-null directions less than unit length. \emph{Neurons:} sextiles of neurons ranked by the norm of their row (or column) in Muon's actual $\NS$ step. Now scale group $b$ by $c_b$, i.e.\ $W_\ell\leftarrow W_\ell-\sum_b c_bD_{\ell,b}$. To second order in $c$, the held-out loss changes by
\begin{equation}\label{eq:quad}
\begin{gathered}
q(c)=-g^\top c+\tfrac12c^\top Hc,\\
g_b=\langle\nabla\hat L,D_b\rangle,\qquad H_{bb'}=\langle D_b,\nabla^2\hat L\,D_{b'}\rangle,
\end{gathered}
\end{equation}
where $\hat L$ is the loss on held-out tokens. Thus $g$ and $H$ refer to the held-out loss, not the training minibatch, though they still carry finite-sample noise. Muon is $c=\mathbf 1$. The stationary point of $q$ is $c^\star=H^{-1}g$. It is the minimizer of $q$ only when $H\succeq0$, and the unique one when $H\succ0$ (\cref{app:lean-quad}). Every curvature matrix we estimate is positive definite, with smallest eigenvalue at least $\minEigAll$. The per-group multiplier $c_b=g_b/H_{bb}$ minimizes $q$ for group $b$ on its own. Muon's flat filter is second-order optimal up to a common scale exactly when $c^\star\propto\mathbf 1$. All $c_b$ scale as $1/\eta$, so only their ratios are comparable across learning rates.

\paragraph{Scale versus shape.} A greedy one-step model is a poor guide to the overall step size, because it ignores the step's effect on future iterates~\citep{shorthorizon,hypergrad}. It can still say which groups get too much or too little of a given step. The following identity separates the two.
\begin{theorem}[Exact scale/shape split; \cref{app:lean-quad}]\label{thm:split}
Let $H$ be symmetric positive definite, $c^\star=H^{-1}g$ the minimizer of $q$, and $\gamma^\star=\mathbf 1^\top g/\mathbf 1^\top H\mathbf 1$ the minimizer of $q(\gamma\mathbf 1)$ over $\gamma$, i.e.\ the best uniform rescaling of Muon's step. Then
\begin{align*}
q(\mathbf 1)-q(c^\star)&=\underbrace{\tfrac12(1-\gamma^\star)^2\,\mathbf 1^\top H\mathbf 1}_{\text{step size}}\\
&\quad+\underbrace{\tfrac12(\gamma^\star\mathbf 1-c^\star)^\top H(\gamma^\star\mathbf 1-c^\star)}_{\text{shape}},
\end{align*}
and the shape term is strictly positive unless $c^\star$ is flat.
\end{theorem}
The identity itself needs only symmetry and $Hc^\star=g$. Positive definiteness is what makes $c^\star$ and $\gamma^\star$ optima, and both terms losses relative to them.
The step-size term carries the known short-horizon bias. Removing it does not make the shape immune: in an explicit example, a one-step shape preference reverses after one more gradient step (\cref{prop:a-horizon}). We therefore treat the shape as a one-step diagnostic whose implications must be tested in training, not as a prescription.

\paragraph{Fixed norm.} \Cref{thm:split} compares Muon with $c^\star$, whose norm generally differs from Muon's. Our experiments instead compare steps of \emph{equal} Frobenius norm. Band directions are mutually orthogonal, with squared norms proportional to the band sizes $|I_b|$, so equal norm means $c^\top Nc=\mathbf 1^\top N\mathbf 1$ with $N=\operatorname{diag}(|I_b|)$. A positive shape term does not by itself imply that such a step beats Muon:
\begin{proposition}[Same-norm reallocation; \cref{app:lean-quad}]\label{prop:fixed}
Let $N\succ0$, $r=H\mathbf 1-g$ the model gradient at Muon's step, and $p$ the component of $r$ orthogonal to $N\mathbf 1$.
(i)~If $p\neq0$, some $c$ with $c^\top Nc=\mathbf 1^\top N\mathbf 1$ has $q(c)<q(\mathbf 1)$.
(ii)~If $r=\lambda N\mathbf 1$ and $H-\lambda N\succeq0$ for some $\lambda$, no such $c$ exists.
(iii)~The shape term does not decide between (i) and (ii). For $H=\operatorname{diag}(1,3)$, $g=(0,2)$ and $N=I$, the shape term is $\tfrac16>0$, yet Muon is optimal among all same-norm steps.
\end{proposition}
We therefore also compute the best same-norm step, an exactly solvable trust-region problem whose solution is certified globally optimal on the ellipsoid (\cref{thm:a-trust}), and the model's prediction for the norm-matched filters we train with. Two remarks make this the right comparison. First, every hidden matrix has rank 768 and bands are defined by rank, so each matrix's step has squared norm proportional to $\sum_b|I_b|c_b^2$; matching each matrix's norm separately, as the experiments do, is the same constraint for exact-polar bands. Second, the ratios $g_b/H_{bb}$ are not the fixed-norm criterion: even with diagonal $H$, the band with the larger ratio can be the one a same-norm step should shrink (\cref{prop:a-ratio}); only $p$, which uses the full $H$, decides.

\paragraph{Estimation and validity.} We take checkpoints of a Muon training run (\cref{sec:experiments}) at steps 1,024 (constant learning rate), 1,536 (start of the cooldown) and 2,048 (end), and form the exact next Nesterov input. Bands are ranks $\{1\text{--}8, 9\text{--}32, 33\text{--}96, 97\text{--}256, 257\text{--}512, 513\text{--}768\}$ of every matrix. $g$, the Hessian--vector products for $H$, and realized loss changes use three disjoint held-out sets (393k, 98k and 393k tokens; \cref{app:diag}). Across 12 test steps (Muon and $c^\star$; bands and neurons; three checkpoints), the predicted and realized changes have the same sign in every case and agree within a factor of three (\cref{fig:calibration}). Most importantly, the $c^\star$ step \emph{realizes} a much larger decrease than Muon's step on unseen tokens: \actStara{} against \actMuona{} at step 1,024. $H$ is ill-conditioned (condition number $\sim\!10^4$), so $c^\star$ is irregular across bands (\cref{app:diag}) and we use it only through realized losses and \cref{thm:split}.

\section{What the anatomy shows}\label{sec:anatomy}

\begin{figure*}[t]
\centering
\includegraphics[width=\textwidth]{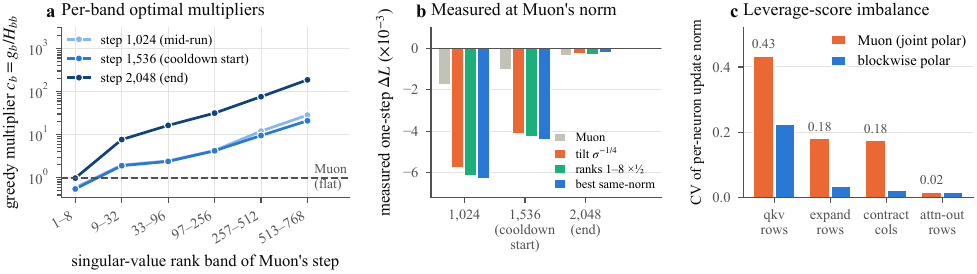}
\caption{\textbf{Anatomy of Muon's step} (124M decoder, Muon seed 43; screen schedule with cooldown over steps 1,536--2,048). \textbf{(a)}~Per-band second-order optimal multiplier $c_b=g_b/H_{bb}$ for the exact-polar step restricted to each singular-value band. Muon's flat filter is $c_b=1$ (dashed). Multipliers scale as $1/\eta$, and $\eta$ is $10\times$ smaller at step 2,048, so compare profiles rather than levels. The tail-to-top ratio is 48, 38 and 190 at the three checkpoints. These isolated-band ratios are not the fixed-norm criterion (\cref{prop:a-ratio}); panel (b) uses the full curvature. \textbf{(b)}~Measured change of the held-out loss after one step at Muon's own update norm (393k tokens; \cref{tab:tiltslope}). Bars show Muon, the exact tilt $\sigma^{-1/4}$, halving ranks 1--8, and the best same-norm filter from the quadratic model (\cref{prop:fixed}). Moving step length off the top helps severalfold mid-training and at the start of the cooldown, and not at its end, where Muon's top band is no longer over-stepped. \textbf{(c)}~Spread (coefficient of variation) of per-neuron update norms of Muon's step, mean over the 12 matrices of each type, at step 1,024. The fused $2304\times768$ QKV matrix carries the largest imbalance. Block-wise polar factors halve the QKV spread and nearly remove the MLP spread.}
\label{fig:anatomy}
\end{figure*}

\paragraph{F1: Muon over-steps its top singular directions.} At every checkpoint the per-band multiplier rises monotonically from the top band to the tail (\cref{fig:anatomy}a). The tail-to-top ratio is 48 at step 1,024, 38 at the start of the cooldown and 190 at the end. Within each of the four layer types separately, the top band has the smallest multiplier in all twelve type--checkpoint cases (\cref{app:diag}). The raw quantities show the mechanism. At step 1,024 the top band's curvature $H_{11}$ is $29\times$ that of the next band, but its gradient alignment $g_1$ is only $9\times$ larger. Curvature falls off faster than signal with rank. Equal step lengths therefore over-step the high-curvature top directions, in which gradients concentrate~\citep{gurari,ghorbani,eos}. The ratios are isolated-band quantities, however. At fixed norm, the full-curvature analysis (F2) confirms the top as the place to take step length from but gives it to ranks 97--512, not to the extreme tail.

\paragraph{F2: at Muon's norm, moving step length off the top helps until late in the cooldown.} At step 1,024 and at the start of the cooldown, only \sharea\% and \shareb\% of the gap of \cref{thm:split} is shape. The rest is the greedy model asking for a smaller step ($\gamma^\star=\gammaa$, \gammab). Even so, most of the gap is reachable \emph{without} changing the step's norm. The model gradient is almost entirely tangent to the same-norm ellipsoid ($\|p\|/\|r\|=\tanga$ and \tangb), so by \cref{prop:fixed} a same-norm step does better. The best one is predicted to change the one-step loss by \fixGaina{} and \fixGainb, against \preMuona{} and \preMuonb{} for Muon; that is \fixSharea\% and \fixShareb\% of the full gap. It takes step length from ranks 1--96 and gives it to ranks 97--512 (\cref{app:diag}). We confirmed this by measuring the actual held-out loss after one step at matched norm (\cref{fig:anatomy}b, \cref{tab:tiltslope}). Relative to Muon, halving ranks 1--8 changes it by \msGainTopa{} and \msGainTopb{} ($\times10^{-3}$), and the best same-norm filter by \msGainFixa{} and \msGainFixb, which is \msRealizeda\% and \msRealizedb\% of the predicted gain. Along the tilt family $\sigma^p$, $dL/dp=\msSlopea$ and \msSlopeb{} at $p=0$, so a negative power helps and a positive one hurts. A proved certificate (\cref{cor:a-chain}) turns a predicted gain into a guaranteed one once the model's errors are bounded; its constants are not yet measured (\cref{app:repro}). As a consistency check, the observed errors, at most $\calMax\times10^{-4}$ at 12 calibration steps and $\msGapDiffa$--$\msGapDiffb\times10^{-3}$ for the same-norm gains, are well below the predicted gains here and far above the one at the end. At the end of the cooldown the picture changes. The top band is no longer over-stepped ($c_1=\cTopc$), the model predicts a gain of only \fixGainc, and the measured one-step losses of all filters agree with Muon's to within $1.2\times10^{-4}$ ($dL/dp\approx0$). \Cref{thm:split} still attributes \sharec\% of the gap to shape there, but that gap is measured against a step of very different norm and is out of reach at Muon's norm. The headroom a norm-matched tilt can use is therefore largest early in the cooldown and gone by its end.

\paragraph{F3: neuron imbalance is concentrated in fused QKV.} Per-neuron update norms of Muon's step vary most in the fused QKV matrix (CV \levQKV, per-layer range \levQKVlo--\levQKVhi). The MLP's expand rows and contract columns have CV \levExp{}, and the square attention-output matrix has CV \levAttn{} (\cref{fig:anatomy}c). Grouping neurons by Muon's update size gives the same monotone picture as the bands: the most-moved sextile has multiplier \nTopa{} and the least-moved \nTaila{} (\cref{app:diag}). Only \nsharea\% of the neuron-group gap at step 1,024 is shape (\nsharec\% at the end), so any benefit of block-wise steps must come mostly from the new directions they produce, not from reweighting.

\paragraph{Implications.} We test three implications in \cref{sec:experiments}.
\textbf{I1:} rebalancing should matter most where the imbalance is largest, in QKV, and little in the MLP.
\textbf{I2:} at fixed norm, moving step length off the top of the spectrum ($p<0$) should help, independently of I1, because it acts on different structure. By F2 its per-step benefit is largest from mid-training to the start of the cooldown and gone by the end, because the decaying learning rate itself removes the top's over-stepping. A tilt confined to the cooldown should therefore build an early lead that partly erodes.
\textbf{I3:} changes that barely alter the allocation should not help. Making $\NS$ a more exact polar map is one: it removes $\NS$'s small deviations from unit length, mostly by lifting near-null directions, and leaves the top's share relative to the bulk essentially unchanged.

\section{From measurements to interventions}\label{sec:interventions}

\paragraph{Neuron allocation: block-wise polar steps.} Following \cref{prop:blocks}, we orthogonalize blocks separately. \emph{Split QKV} uses three square $768\times768$ blocks, one each for Q, K and V. \emph{Per-head} uses 36 wide $64\times768$ blocks, whose polar factors have exactly orthonormal rows. \emph{Square blocks} splits QKV and also splits the MLP's $3072\times768$ and $768\times3072$ matrices into four square blocks each. Wide matrices' column blocks are scaled by $1/\sqrt k$ to keep Muon's update energy. All reported block arms are norm-matched anyway; the factor matters for the deployable version (\cref{app:impl}). The stateful alternative from prior work is NorMuon's row normalization by an EMA of row second moments~\citep{normuon}.

\paragraph{Spectral allocation: a matmul-only cooldown tilt.} The exact tilt $U\,\phi_5(\Sigma)(\Sigma/\sigma_{\rm rms})^pV^\top$ needs an eigendecomposition per matrix and step. We avoid it as follows. Let $O=\NS(N)=U\operatorname{diag}(f_i)V^\top$ with $f_i=\phi_5(\hat\sigma_i)$, and $S=ON^\top=U\operatorname{diag}(f_i\sigma_i)U^\top\succeq0$. A polynomial in $S$ acts in the same singular basis (\cref{prop:gram}):
\[
D=P(S/\Lambda)\,O=U\operatorname{diag}\big(f_i\,P(f_i\sigma_i/\Lambda)\big)V^\top .
\]
With $P(x)\approx x^p$ the filter is $f_i^{1+p}\sigma_i^p$ up to scale. On the Newton--Schulz plateau ($f_i\in[0.68,1.20]$) this is the exact tilt $f_i\sigma_i^p$ up to a factor in $[0.96,1.10]$; near-null directions are lifted further. We use a degree-12 least-squares Chebyshev fit of $x^{p}$ on $[10^{-3},1]$. $\Lambda$ is $1.01$ times a 20-step power-iteration estimate of $\lambda_{\max}(S)$, which is an estimate rather than a proved bound. We then rescale $D$ to $\|O\|_\F$ (Algorithm~\ref{alg:step}).
\begin{proposition}[What the tilt computes; \cref{app:lean-ns}]\label{prop:gram}
For any coefficients and depth $k$ and any real polynomial $P$, let $O=\varphi^k(X)$ and $D=P(OX^\top)\,O$. (i)~$D$ has the form $F(XX^\top)X$, so $D(XQ)=D(X)Q$ for every orthogonal $Q$. (ii)~If $X=U\operatorname{diag}(d)V^\top$ with orthonormal columns, then $O=U\operatorname{diag}(f_i)V^\top$ and $D=U\operatorname{diag}(f_iP(f_id_i))V^\top$, with $f_i$ the scalar iterate of $d_i$.
\end{proposition}
Part (ii), not (i), is what keeps the singular vectors: $X\mapsto AX$ is also a Gram map. The normalization, $\Lambda$ and the final rescaling are scalars. The tilted step therefore changes \emph{only} how singular directions are weighted, so a comparison with Muon isolates the spectral allocation. One numerical detail decides whether the method works. Evaluating $P$ in monomials by Horner's rule in TF32 cancels catastrophically (coefficients reach $4\times10^4$) and flipped the filter's sign on the top direction. With square blocks as the base (\scrBlocks{} alone, against fused-QKV Muon), that version reached \scrPolyBad, against \scrPolyGood{} for Clenshaw's recurrence~\citep{clenshaw,trefethen} in FP32 and \scrBlocksExact{} for the exact tilt; the Clenshaw filter matches the exact tilt to direction cosine 0.998--0.999, so the normalized steps differ by at most $6.3\%$ in Frobenius norm.

\section{Experiments}\label{sec:experiments}

\paragraph{Setup.} We train a 124.5M-parameter pre-norm decoder (12 layers, width 768, 12 heads, tied embeddings) on a 167.8M-token FineWeb-Edu sample~\citep{fineweb} with the GPT-2 tokenizer, sequence length 512 and 24,576 tokens per step. Muon~\eqref{eq:muon} acts on the 48 hidden matrices with momentum 0.95 and decoupled weight decay 0.1. Its state is in FP32, and Newton--Schulz runs in FP32 with TF32 matrix products. AdamW~\citep{adamw} ($3\times10^{-4}$) handles embeddings and norms, and gradients are clipped at norm 1.0. The schedule is warmup--constant--cooldown~\citep{wsd}: 128 warmup steps, then linear decay to $0.1\times$ over the last 25\% of steps. The \emph{screen} runs 2,048 steps (50.3M tokens). The \emph{confirmation} runs 4,096 steps (100.7M tokens). The metric is final loss on 4.18M held-out tokens. A 355M model (24 layers, width 1024, 16 heads) tests scale.

\paragraph{Controls.} \textbf{Norm matching:} unless stated otherwise, every direction-changing arm is rescaled, per matrix and per step, to the Frobenius norm of Muon's $\NS$ output on the same momentum, so differences reflect direction, not step size. \textbf{Tuned baselines:} on the screen, Muon's learning rate was swept over $\{0.0035,0.005,0.007,0.01\}$, split-QKV Muon's over $\{0.0035,0.005,0.007\}$, and the 355M baseline's over the same three values. 0.005 is an interior optimum in all three cases, and every new arm uses the baseline's rate. \textbf{Paired seeds:} each seed fixes initialization and data order. Confirmations use seeds 42--44 for every arm, and we report paired 95\% $t$-intervals. \textbf{A/A check:} before the cooldown the tilt arms run an optimizer mathematically identical to their baselines. Their intermediate evaluations differ by at most \aaMax\nats, an empirical bound on numerical non-determinism plus evaluation noise. The screen baseline's seed spread is \scrMuonRange\nats. \textbf{Gating:} midway through the study we adopted a rule that a full-length run is launched only if its screen beats the baseline by more than 0.005\nats. The factorial ablations and the 355M seeds 43--44 were released under it; the earlier confirmations were not.

\begin{table}[t]
\centering
\caption{\textbf{Confirmation at 100.7M tokens} (124M model, three paired seeds per arm). Mean final held-out loss and mean paired differences against the original fused-QKV Muon and against the stronger split-QKV Muon. Brackets give 95\% $t$-intervals for the latter; intervals against fused-QKV Muon are in the text. Lower is better.}
\label{tab:main}
\small\setlength{\tabcolsep}{3.5pt}
\resizebox{\columnwidth}{!}{\begin{tabular}{@{}lccc@{}}
\toprule
Optimizer & Loss & $\Delta$ vs.\ Muon & $\Delta$ vs.\ split-QKV \\
\midrule
Muon (fused QKV) & 3.9427 & --- & $+$0.0250 \\
NorMuon (fused QKV) & 3.9277 & $-$0.0150 & $+$0.0100\,{\tiny[$+$0.003, $+$0.017]} \\
Square blocks & 3.9187 & $-$0.0239 & $+$0.0011\,{\tiny[$-$0.003, $+$0.005]} \\
\midrule
Split-QKV Muon & 3.9176 & $-$0.0250 & --- \\
\quad + cooldown tilt & 3.9057 & $-$0.0370 & $-$0.0120\,{\tiny[$-$0.014, $-$0.010]} \\
\quad + NorMuon & 3.9139 & $-$0.0288 & $-$0.0038\,{\tiny[$-$0.006, $-$0.001]} \\
\textbf{\quad + NorMuon + tilt} & 3.9018 & $-$0.0408 & $-$0.0158\,{\tiny[$-$0.018, $-$0.014]} \\
\quad + NorMuon + tilt + Schatten-1 decay & 3.8966 & $-$0.0460 & $-$0.0210\,{\tiny[$-$0.024, $-$0.018]} \\
\midrule
Sq.\ blocks + NorMuon + tilt & 3.9050 & $-$0.0376 & $-$0.0126\,{\tiny[$-$0.015, $-$0.010]} \\
\bottomrule
\end{tabular}
}
\end{table}

\begin{figure*}[t]
\centering
\includegraphics[width=\textwidth]{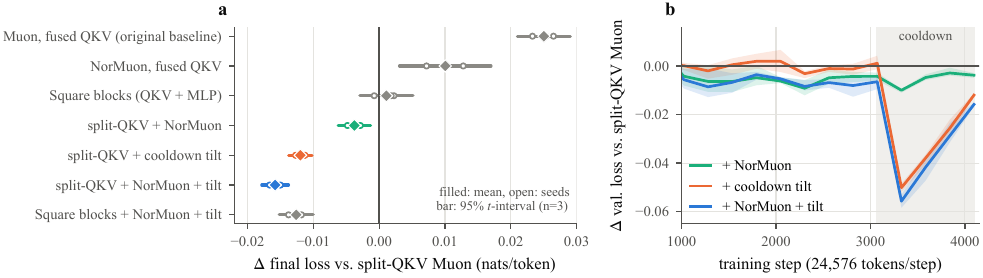}
\caption{\textbf{Controlled results at 100.7M tokens} (124M, three paired seeds). \textbf{(a)}~Paired difference in final loss against split-QKV Muon. Splitting Q, K and V explains the whole square-block gain; NorMuon on fused QKV is worse than the split; the tilt and NorMuon add on top of the split. \textbf{(b)}~Paired difference over training (mean, with the seed range shaded). Intermediate points use 98k held-out tokens; the last point uses 4.18M. Before the cooldown the tilt arm is mathematically identical to its baseline (A/A differences $\le$\aaMax). The tilt's lead peaks at \tiltPeak\nats{} 256 steps into the cooldown and settles at \dSTvS.}
\label{fig:results}
\end{figure*}

\paragraph{I1: the neuron gain is the QKV split.} Splitting Q, K and V changes the final loss by \dSvM\nats{} \ciSvM{} relative to fused-QKV Muon (\cref{tab:main}). It is better in every seed, and fused-QKV Muon would need about \tokSvM\% more tokens to match. Square blocks everywhere add nothing beyond it (\dBvS{} \ciBvS). Square blocks in the MLP alone, with QKV fused, change the loss by \scrMLPBlocks{} (screen). As I1 implies, the large QKV imbalance is the costly one and the moderate MLP imbalance is harmless. The data also locate the effect at the level of projections rather than rows. Per-head blocks equalize rows exactly yet do worse than three blocks (\scrHeads{} against \scrSplit, screen). NorMuon's row normalization on fused QKV is worse than the split (\dNvS{} \ciNvS), and on top of the split it adds only \dSNvS{} \ciSNvS. One reading is that in a fused matrix Q, K and V share a single spectral budget, so the projection with the strongest momentum claims most of it, whereas splitting gives each projection its own budget. We have not isolated this mechanism. Either way, methods evaluated against fused-QKV Muon can take credit for a gain the split provides at no cost.

\paragraph{I2: the tilt helps, and adds to neuron balance.} On the split baseline the cooldown tilt $\sigma^{-1/4}$ changes the loss by \dSTvS\nats{} \ciSTvS{} and is better in all three seeds (\psSTvS). The $2{\times}2$ factorial over \{NorMuon, tilt\} on split-QKV Muon is additive within noise: \dSNvS{} and \dSTvS{} separately, \dSNTvS{} \ciSNTvS{} together. Each component's effect is the same with or without the other: the tilt given NorMuon gives \dSNTvSN, and NorMuon given the tilt gives \dSNTvST. The tilt supplies about three quarters of the combined gain. By the split baseline's own loss-versus-tokens curve (\slopeS\nats{} per doubling, a two-point log-linear estimate), split-QKV Muon would need about \tokSNTvS\% more tokens to match the combination. Other tilts and band filters behave consistently on the screen (\cref{app:screen}); the dose was not optimized. Consistent with F2, timing and dose matter little: on the full stack, starting the tilt at step 1,024 or using $\sigma^{-1/2}$ changes the screen loss by only \scrFullEarly{} and \scrFullHalf{} (one seed each, within the 0.003 seed spread). We keep the tilt in the cooldown, where it costs least.

\paragraph{How the gain develops.} \Cref{fig:results}b shows the tilt's lead peaking at \tiltPeak\nats{} 256 steps into the cooldown and narrowing to \dSTvS{} at the end. This is the pattern F2 predicts. At fixed norm the per-step benefit of moving step length off the top is large at the start of the cooldown and gone by its end, because the decaying learning rate itself removes the top's over-stepping, so the baseline recovers part of the early lead. The baseline's cooldown length and final learning rate were not tuned, so a different cooldown might recover part of the final gain; we have not tested this.

\paragraph{Scale.} At 355M parameters (GPT-2-medium shape, 96 Muon matrices, same data and tokens), the split-QKV baseline's learning-rate screen again peaks at 0.005 (\bigLRa/\bigLRb/\bigLRc{} for 0.0035/0.005/0.007). Against the baseline at the same seeds, NorMuon $+$ tilt changes the final loss by \dBig\nats{} \ciBig{} (per seed \psBig), comparable to the 124M effect; the baseline's \bigSn{} seeds span only \bigSspread\nats.

\paragraph{I3 and other geometric changes.} \Cref{tab:theory} (\cref{app:screen}) collects them. Three extra cubic Newton--Schulz iterations drive every singular value of the $\NS$ output to 1 and change the loss by \scrPolish, within noise, as I3 implies. \emph{Input whitening}, steepest descent in the data metric $\|DC^{1/2}\|_\op$, has the cleanest theory of anything we tried. In its exact form it is $\mathrm{GL}(n)$-equivariant and conditioning-free on least squares (\cref{app:lean-white}); the damped version we run is approximately so. It is harmful: $+0.021$ to $+0.060$ when norm-matched, and \scrWhNat{} at its own magnitude. A whitened step concentrates its energy in low-variance input directions, so at a fixed norm it moves the layer's outputs less. The published Toner recipe~\citep{toner} gives \scrToner{} (one seed), the sign of its reported gain at smaller size. For \emph{decay geometry}, energy-matched Schatten-4 decay, which mainly shrinks the top of the spectrum, hurts (\scrSchFour). Schatten-1 decay, which removes about the same amount from every singular value and so prunes the bulk, helps (\scrSchOneMean{} \scrSchOneCI, three seeds). Decay along the polar factor of $W$ is prior art, used with a fixed coefficient for low rank, where with Muon it costs loss~\citep{swd}, and for grokking~\citep{lowrankdecay}. With our energy-matched coefficient $\|W\|_\F^2/\|W\|_*$ (\cref{app:lean-decay}) it is additive with our method: \dSNTDvSNT{} \ciSNTDvSNT{} on top of split QKV $+$ NorMuon $+$ tilt at 100.7M tokens (three seeds, \cref{tab:main}), \dSNTDvS{} against split-QKV Muon, at \timeSNTDvS\% more wall time. In this regime the useful role of decay is to clean the bulk, not to cap the spectral norm.

\paragraph{Cost.} Splitting QKV adds no FLOPs; its \timeSvM\% recorded wall time is mostly the norm-matching scaffold. In our unfused FP32 implementation NorMuon $+$ tilt takes \timeSNTvS\% more wall time than split-QKV Muon at 124M and \timeBig\% at 355M, against a token saving of about \tokSNTvS\%. \textbf{The gain is in tokens, not yet in wall clock;} folding the tilt into the Newton--Schulz coefficients, as DynMuon does, may remove most of the overhead.

\section{What formal verification did and did not buy}\label{sec:lean}

All formal results are stated in \cref{app:lean}. The Newton--Schulz results hold for the iteration Muon actually runs, with any coefficients, depth and normalizer. The block results concern exactly orthogonal blocks.
\textbf{What it bought.}
(i)~\emph{Bookkeeping.} The energy identity (\cref{prop:blocks}), carried over to wide matrices, gives the $1/\sqrt k$ factor without which a deployable block step on the MLP contraction is $1.9\times$ too large.
(ii)~\emph{A symmetry guarantee.} The cheap tilt keeps Muon's right equivariance (\cref{prop:gram}).
(iii)~\emph{A correction to our own reasoning.} \Cref{thm:split}'s stationary points are optima only under positive definite curvature, and its shape term compares Muon with a step of a different norm. Formalizing the question our norm-matched experiments ask produced \cref{prop:fixed} and a finite-step certificate, and changed our reading of the diagnostic (F2).
(iv)~\emph{A change of direction.} Earlier counterexamples (\cref{app:lean}) showed that spatial orientation and curvature, not temporal statistics, decide momentum preferences under $\NS$.
\textbf{What it did not buy} was a prediction of training benefit. Whitened Muon is provably equivariant and conditioning-free yet hurts; the joint polar factor is provably suboptimal for the block norm, yet MLP blocks do not help. Whether a certified property is the bottleneck is an empirical question.

\section{Related work}\label{sec:related}

\textbf{Muon and its geometry.} Muon~\citep{muon} and its scaling studies~\citep{moonlight,essentialai} have been analyzed as steepest descent under operator norms~\citep{bernstein,modularnorm}, as linear-minimization-oracle methods~\citep{scion,gluon}, as non-Euclidean trust regions~\citep{kovalev} and as constrained optimization~\citep{chenlili,muonconv}. Prior work also studies polar approximations~\citep{polarexpress} and the benefit of flat spectra for tail-end memories~\citep{tailend}. On matrix quadratics, \citet{quadlens} show that the polar direction beats the gradient direction, each at its best step size, when large singular directions sit in high curvature, and that finite Newton--Schulz acts as a damped polar update.

\textbf{Spectra, neurons and blocks.} DynMuon schedules a spectral power from $+1$ to $-\tfrac14$~\citep{dynmuon}. Freon/Kaon show that random or inverted spectra can work as well as flat ones~\citep{freon}, and Spectrally Targeted Muon, AdaMuon and PolarGrad adapt magnitudes~\citep{stmuon,adamuon,polargrad}. NorMuon and Muon+ normalize rows~\citep{normuon,muonplus}, Aurora gives the leverage-score argument~\citep{aurora}, and row-sphere and Stiefel constraints are compared by \citet{rowsphere}. MuonBP and HiMuon orthogonalize blocks~\citep{muonbp,himuon}, and separate Q/K/V orthogonalization is standard in modded-nanogpt~\citep{moddednanogpt}. Closest to our diagnostic, SAMuon measures held-out optimal step sizes along each singular direction of the momentum, finding a volatile head and a tolerant bulk, and builds an optimizer from the profile~\citep{samuon}. BulkBoost proves when a norm-matched bulk/spike reweighting improves an objective to first order~\citep{bulkboost}, and \citet{wangcurv} attribute Muon's advantage over Adam to within- and cross-layer curvature of its update. Our tilt uses DynMuon's dose and late-training sign. We add the full cross-group curvature, neuron groups, scale versus fixed-norm shape with proved criteria, and a controlled attribution: the QKV split accounts for the block-wise gains and most of NorMuon's gain over fused-QKV Muon.

\textbf{Data metric, curvature and evaluation.} Shampoo, SOAP and K-FAC precondition with layer statistics~\citep{shampoo,soap,kfac}, and whitened Muon has close 2026 relatives~\citep{toner,gomuon,mousse,newtonmuon,fismo}. Gradients concentrate in a few top Hessian directions~\citep{gurari} above a bulk~\citep{sagun,ghorbani,papyan}, near the edge of stability~\citep{eos}. Greedy step sizes suffer short-horizon bias~\citep{shorthorizon,hypergrad}, and optimizer gains shrink against tuned baselines~\citep{fantastic,deconstructing}. Like \citet{samuon} and \citet{wangcurv}, we measure curvature along the optimizer's own step, here with held-out Hessian--vector products~\citep{pearlmutter}. Formal verification of ML systems goes back at least to \citet{selsam}.

\textbf{Novelty.} Held-out spectral probing of Muon's step~\citep{samuon} and first-order analysis of norm-matched reallocation~\citep{bulkboost} precede this work. We found no prior work that measures the full cross-group curvature of the step by singular band and by neuron group, separates scale from fixed-norm shape with proved criteria and counterexamples, and uses the result in a controlled attribution of published ingredients.

\section{Limitations and conclusion}\label{sec:conclusion}

\textbf{Scale and regime.} One data distribution; 124M and 355M parameters (three seeds each); at most 100.7M tokens, under one token per parameter. Effect sizes are specific to this regime. \textbf{The diagnostic is local and noisy.} It is second-order, one step ahead and from a single run, without intervals. $g$ and $H$ come from separate held-out folds, and its joint optimum is ill-conditioned. It perturbs only the hidden matrices along exact-polar band directions, leaving out the AdamW and decay parts of the update. The measured one-step losses use one checkpoint seed and one validation subset, without intervals, and the certificate's constants are not yet measured. We used the diagnostic for direction, not dose; the tilt power comes from prior work. The fixed-norm analysis covers bands only. \textbf{Design and statistics.} The anatomy was not pre-registered; intervals are nominal (three paired seeds); most screen arms are single-seed. \textbf{Alternatives.} The baseline's cooldown was not tuned; the QKV mechanism is a hypothesis. \textbf{Provenance.} Comparisons with fused-QKV or square-block arms cross code versions; split-QKV arms share one (\cref{app:repro}).

\textbf{Conclusion.} Measured against curvature, both of Muon's hidden allocations are miscalibrated. Its spectrum over-steps the top, which a same-norm reallocation can exploit until late in the cooldown, and its per-neuron steps are most uneven in fused QKV matrices. Corrections aimed at these help additively; principled changes that miss them do not.

\bibliographystyle{plainnat}
\bibliography{references}

\clearpage
\appendix
\onecolumn

\section{Formal results}\label{app:lean}

This appendix states, in ordinary mathematical language, every formal result the paper uses. All of them are proved in Lean~4 with Mathlib, in exact real arithmetic. They certify the algebra and geometry behind each method. They say nothing about floating-point behaviour, convergence or training benefit.

\paragraph{Notation.} $\langle A,B\rangle=\operatorname{tr}(A^\top B)$. For real coefficients $a,b,c$ let $\varphi(X)=aX+b\,XX^\top X+c\,(XX^\top)^2X$, and let $\varphi^{k}$ be its $k$-fold composition. For any function $\nu$ of an $m\times m$ matrix, the \emph{Newton--Schulz map} is
\[
\mathrm{NS}(X)=\varphi^{k}\big(\nu(XX^\top)\,X\big).
\]
Muon's $\NS$ is the case $k=5$, $(a,b,c)=(3.4445,-4.775,2.0315)$, $\nu(XX^\top)=1/\sqrt{\operatorname{tr}XX^\top}$. Results stated for $\mathrm{NS}$ hold for \emph{every} choice of $a,b,c$, $k$ and $\nu$, and so apply to the iteration Muon actually runs, not only to an exact polar factor. A matrix $Q$ is orthogonal if $QQ^\top=I$.

\subsection{Newton--Schulz maps and the matmul-only tilt}\label{app:lean-ns}

\begin{adefinition}[Gram map]
A map $G:\mathbb R^{m\times n}\to\mathbb R^{m\times n}$ is a \emph{Gram map} if there is a function $F:\mathbb R^{m\times m}\to\mathbb R^{m\times m}$ such that $G(X)=F(XX^\top)\,X$ for every $X$.
\end{adefinition}

\begin{atheorem}[Gram maps commute with right rotations]\label{thm:a-gram}
If $G$ is a Gram map and $Q\in\mathbb R^{n\times n}$ is orthogonal, then $G(XQ)=G(X)\,Q$ for every $X$.
\end{atheorem}
\emph{Proof idea.} $(XQ)(XQ)^\top=XX^\top$, so $G(XQ)=F(XX^\top)XQ$.

\begin{atheorem}[Newton--Schulz iterates are Gram maps]\label{thm:a-ns}
For all $a,b,c$ and $k$, the map $X\mapsto\varphi^{k}(X)$ is a Gram map. Consequently $\mathrm{NS}(XQ)=\mathrm{NS}(X)\,Q$ for every orthogonal $Q$, for any normalizer $\nu$.
\end{atheorem}
\emph{Proof idea.} Each step multiplies $X$ on the left by the matrix $aI+b\,XX^\top+c\,(XX^\top)^2$. That matrix depends only on the Gram matrix, and the class of Gram maps is closed under composition.

\begin{atheorem}[The polynomial tilt preserves Muon's symmetry]\label{thm:a-tilt}
Let $P$ be any real polynomial and $O=\varphi^{k}(X)$. The tilt $T_P(X)=P(OX^\top)\,O$ is a Gram map. Hence $T_P(XQ)=T_P(X)\,Q$ for every orthogonal $Q$.
\end{atheorem}
Right equivariance alone does not mean singular vectors are kept: $X\mapsto AX$ for a fixed non-scalar $A$ is also a Gram map. The next theorem shows that the Newton--Schulz iterates and the tilt do keep them.

\begin{atheorem}[Singular-basis action of Newton--Schulz and of the tilt]\label{thm:a-svd}
Let $X=U\operatorname{diag}(d)V^\top$, where $U$ and $V$ have orthonormal columns ($U^\top U=I$, $V^\top V=I$; any shapes, e.g.\ a thin SVD). Let $s_k$ be the scalar map $x\mapsto ax+bx^3+cx^5$ iterated $k$ times, and $f_i=s_k(d_i)$.
(i)~$\varphi^k(X)=U\operatorname{diag}(f_i)V^\top$.
(ii)~For every real polynomial $P$, $P(\varphi^k(X)X^\top)\,\varphi^k(X)=U\operatorname{diag}\big(f_i\,P(f_i d_i)\big)V^\top$.
\end{atheorem}
\emph{Proof idea.} $XX^\top=U\operatorname{diag}(d_i^2)U^\top$ and $U^\top U=I$ make every product collapse onto the diagonal. Induct on $k$ for (i); for (ii), induct on the powers of $S$, then use linearity in $P$.

In the implementation $X$ is first normalized and $S$ is divided by an estimate $\Lambda$ of its largest eigenvalue. Both are scalar factors, absorbed into $d$ and $P$. Both singular frames are therefore kept, and the implemented filter is $f_i\,P(f_i\sigma_i/\Lambda)$, not $f_i\,P(\sigma_i)$. With $P(x)\approx x^p$ it is $f_i^{1+p}\sigma_i^p$ up to scale. On the Newton--Schulz plateau ($f_i\in[0.68,1.20]$) this differs from the exact tilt $f_i\sigma_i^p$ by the factor $f_i^{p}$, which lies in $[0.96,1.10]$ for $p=-\tfrac14$.

\begin{atheorem}[From filter error to step error]\label{thm:a-transfer-err}
(i)~For non-zero $Y,\hat Y$ in a real normed space and $\rho\ge0$, $\big\|\rho\hat Y/\|\hat Y\|-\rho Y/\|Y\|\big\|\le2\rho\,\|\hat Y-Y\|/\|Y\|$.
(ii)~For unit vectors $u,v$ in a real inner-product space, $\|u-v\|^2=2(1-\langle u,v\rangle)$.
\end{atheorem}
By (i), a raw filter error of $\epsilon\|Y\|$ moves the norm-matched step by at most $2\epsilon$ of its norm. By (ii), the measured direction cosine of 0.998 between the matmul and exact tilts means their normalized steps are within $6.3\%$.

\subsection{Block geometry and per-neuron step sizes}\label{app:lean-blocks}

\begin{adefinition}[Steepest direction]
For a norm-like function $\mathcal N$ and a gradient $G$, a matrix $D$ is a \emph{steepest direction} if $\mathcal N(D)\le1$ and $\langle G,D'\rangle\le\langle G,D\rangle$ for every $D'$ with $\mathcal N(D')\le1$.
\end{adefinition}

\begin{atheorem}[Block-wise steepest descent]\label{thm:a-blocks}
Let $G=(G_1,\dots,G_p)$ be a block gradient and $\mathcal N_1,\dots,\mathcal N_p$ block norms. If each $D_i$ is a steepest direction for $\mathcal N_i$ at $G_i$, then $(D_1,\dots,D_p)$ is a steepest direction at $G$ for the max-norm $\|D\|_{\max}=\max_i\mathcal N_i(D_i)$, whose unit ball is the product of the block unit balls. With every $\mathcal N_i$ the spectral norm, the block-wise polar factors are therefore steepest for $\|\cdot\|_{\rm blk}$.
\end{atheorem}

\begin{aproposition}[The joint polar step can be strictly worse]\label{prop:a-joint}
Take two $1\times2$ blocks $G_1=G_2=(1,0)$. The polar factor of the stacked matrix has blocks $D_1=D_2=(1/\sqrt2,\,0)$, a partial isometry with $D_1^\top D_1+D_2^\top D_2=e_1e_1^\top$, and attains $\langle G,D\rangle=\sqrt2$. The block-wise step $D_1=D_2=(1,0)$ is feasible for the max-norm and attains $2>\sqrt2$.
\end{aproposition}

\begin{aproposition}[Row norms of polar factors]\label{prop:a-rows}
(i)~If $O$ is square and $OO^\top=I$, every row of $O$ has unit norm.
(ii)~If $O\in\mathbb R^{m\times k}$ has orthonormal columns ($O^\top O=I_k$), its squared row norms sum to $k$.
(iii)~They need not be equal: $O=(1;\,0)\in\mathbb R^{2\times1}$ satisfies $O^\top O=1$ and has row norms 1 and 0.
\end{aproposition}
The squared row norms in (ii) are the leverage scores of $\operatorname{range}(O)$. In Muon's step they are the per-neuron step sizes of a tall matrix.

\begin{aproposition}[Energy identity for block-wise steps]\label{prop:a-energy}
Let $O\in\mathbb R^{m\times k}$ with $O^\top O=I_k$, let $B_1,\dots,B_p\in\mathbb R^{k\times k}$ be orthogonal, and let $c^2=p$. Then
\[
\|c\,O\|_\F^2=\sum_{b=1}^p\|B_b\|_\F^2\;(=pk).
\]
For a tall matrix with $m=pk$ rows, Muon's shape factor $s=\sqrt{m/k}=\sqrt p$ therefore gives the joint step the same Frobenius energy as $p$ orthogonal $k\times k$ blocks, and the learning rate transfers.
\end{aproposition}

\subsection{Second-order model of a filtered step}\label{app:lean-quad}

For group multipliers $c\in\mathbb R^B$, a gradient-alignment vector $g$ and a symmetric curvature matrix $H$, let $q(c)=-g^\top c+\tfrac12c^\top Hc$ (\cref{eq:quad}).

\begin{atheorem}[Gap formula]\label{thm:a-gap}
If $H$ is symmetric and $Hc^\star=g$, then for every $c$,
\[
q(c)-q(c^\star)=\tfrac12\,(c-c^\star)^\top H\,(c-c^\star).
\]
\end{atheorem}

\begin{atheorem}[Scale/shape split]\label{thm:a-split}
Let $H$ be symmetric with $Hc^\star=g$, let $u$ be a direction (Muon's step is $u=\mathbf 1$), and let $\gamma$ satisfy $u^\top H(\gamma u-c^\star)=0$. Then
\[
q(u)-q(c^\star)=\tfrac12(1-\gamma)^2\,u^\top Hu\;+\;\tfrac12\,(\gamma u-c^\star)^\top H\,(\gamma u-c^\star).
\]
When $H$ is positive definite (\cref{thm:a-min}), the first term is the loss due to the overall step size along $u$ and the second the loss due to the shape of $u$.
\end{atheorem}
\emph{Proof idea.} Write $u-c^\star=(1-\gamma)u+(\gamma u-c^\star)$ and apply \cref{thm:a-gap}. The cross term vanishes by the condition on $\gamma$, so the identity is Pythagoras in the $H$-inner product.

\begin{alemma}[The best uniform rescaling is an $H$-projection]\label{lem:a-gamma}
If $u^\top Hu\ne0$, then $\gamma^\star=u^\top g/u^\top Hu$ satisfies $u^\top H(\gamma^\star u-c^\star)=0$. So \cref{thm:a-split} applies with $\gamma=\gamma^\star$, the best scalar multiple of $u$.
\end{alemma}

\begin{alemma}[The shape term is strictly positive]\label{lem:a-pos}
If $H$ is positive definite and $\gamma u\neq c^\star$, then $\tfrac12(\gamma u-c^\star)^\top H(\gamma u-c^\star)>0$. Muon's flat step therefore loses something to shape unless the second-order optimum is itself flat.
\end{alemma}

\paragraph{When the stationary points are optima.} \Cref{thm:a-gap,thm:a-split} are identities that hold for any symmetric $H$. Calling $c^\star$ and $\gamma^\star$ optima, and the two terms of \cref{thm:a-split} losses, needs curvature assumptions:

\begin{atheorem}[Stationary points are minimizers under positive curvature]\label{thm:a-min}
Let $H$ be symmetric and $Hc^\star=g$.
(i)~If $H$ is positive semidefinite, then $q(c^\star)\le q(c)$ for every $c$.
(ii)~If $H$ is positive definite, then $q(c^\star)<q(c)$ for every $c\neq c^\star$.
\end{atheorem}

\begin{alemma}[The best uniform rescaling minimizes along the ray]\label{lem:a-ray}
If $u^\top Hu>0$, then $q(\gamma^\star u)\le q(\gamma u)$ for every $\gamma$, where $\gamma^\star=u^\top g/u^\top Hu$. Indeed $q(\gamma u)-q(\gamma^\star u)=\tfrac12\,u^\top Hu\,(\gamma-\gamma^\star)^2$.
\end{alemma}

\paragraph{Comparison at a fixed update norm.} The shape term of \cref{thm:a-split} compares $u$ with $c^\star$, which generally has a different norm. Experiments that norm-match every arm compare steps on the ellipsoid $\{c:\ c^\top Nc=u^\top Nu\}$ for a symmetric positive definite $N$. Write $r=Hu-g$ for the gradient of $q$ at $u$, $w=Nu$, and
\[
p=r-\frac{r^\top w}{w^\top w}\,w
\]
for the component of $r$ orthogonal to $w$.

\begin{atheorem}[Lagrange identity on the same-norm ellipsoid]\label{thm:a-lagrange}
Let $H$ and $N$ be symmetric. For every $c$ with $c^\top Nc=u^\top Nu$ and every real $\lambda$,
\[
q(c)-q(u)=(r-\lambda Nu)^\top(c-u)+\tfrac12\,(c-u)^\top(H-\lambda N)(c-u).
\]
\end{atheorem}
\emph{Proof idea.} Expand $q(u+d)-q(u)$, and subtract $\tfrac\lambda2$ times the constraint, which in terms of $d=c-u$ reads $2u^\top Nd+d^\top Nd=0$.

\begin{acorollary}[Certificate that no same-norm step improves]\label{cor:a-cert}
If $r=\lambda Nu$ and $H-\lambda N$ is positive semidefinite for some $\lambda$, then $q(u)\le q(c)$ for every $c$ with $c^\top Nc=u^\top Nu$.
\end{acorollary}

\begin{atheorem}[A same-norm improvement exists when the tangential gradient is non-zero]\label{thm:a-improve}
Let $N$ be symmetric positive definite, $u\neq0$, and $p\neq0$. Then there is $c$ with $c^\top Nc=u^\top Nu$ and $q(c)<q(u)$.
\end{atheorem}
\emph{Proof idea.} The vector $v=-p$ satisfies $w^\top v=0$ and $r^\top v=-\|p\|^2<0$. Rescale it to $e=\kappa v$ with $e^\top Ne=u^\top Nu$; then $u^\top Ne=0$. Every $c=xu+ye$ with $x^2+y^2=1$ lies on the ellipsoid. Along the rational parametrization $x=(1-s^2)/(1+s^2)$, $y=2s/(1+s^2)$,
\[
q(c)-q(u)=\frac{2s\big[\,r^\top e+s\,\alpha_1+s^2\alpha_2+s^3\alpha_3\,\big]}{(1+s^2)^2}
\]
for constants $\alpha_i$. Choosing $s=\min\{1,\,-r^\top e/(2K)\}$ with $K=|\alpha_1|+|\alpha_2|+|\alpha_3|+1$ makes the bracket at most $r^\top e/2<0$.

\begin{aproposition}[A positive shape term does not imply a same-norm improvement]\label{prop:a-counter}
Let $H=\operatorname{diag}(1,3)$, $g=(0,2)$, $u=(1,1)$ and $N=I$. The minimizer is $c^\star=(0,\tfrac23)$, the best uniform rescaling is $\gamma^\star=\tfrac12$, and the shape term of \cref{thm:a-split} equals $\tfrac16>0$. Yet $q(u)\le q(c)$ for every $c$ with $\|c\|=\|u\|$. Indeed, on that circle $q(c)-q(u)=(c_2-1)^2$.
\end{aproposition}
Here $r=(1,1)=1\cdot Nu$ and $H-N=\operatorname{diag}(0,2)\succeq0$, so \cref{cor:a-cert} applies. \Cref{thm:a-improve,cor:a-cert} together give the decision rule used in \cref{sec:anatomy}: compute $p$, and if it is non-zero, solve for the best same-norm step.

\paragraph{Certifying the best same-norm step and a finite-step gain.}

\begin{atheorem}[Certificate for the best same-norm step]\label{thm:a-trust}
Let $H$ and $N$ be symmetric. If $c$ satisfies $c^\top Nc=R$, $(H+\lambda N)c=g$ and $H+\lambda N\succeq0$, then $q(c)\le q(c')$ for every $c'$ with $c'^\top Nc'=R$.
\end{atheorem}
The best same-norm steps reported in \cref{sec:anatomy} satisfy these conditions at all three checkpoints: $\lambda>0$, so $H+\lambda N\succ0$, and the stationarity residual is below $10^{-15}$. They are therefore the global optima of the model on the ellipsoid.

\begin{alemma}[Model-error budget]\label{lem:a-budget}
For estimates $(\hat g,\hat H)$ of $(g,H)$ and any $c$, $|\hat q(c)-q(c)|\le|(\hat g-g)^\top c|+\tfrac12|c^\top(\hat H-H)c|$.
\end{alemma}
With $\|\hat g-g\|\le\delta_g$ and $\|\hat H-H\|_\op\le\delta_H$ this is at most $\delta_g\|c\|+\tfrac12\delta_H\|c\|^2$.

\begin{atheorem}[A-posteriori finite-step certificate]\label{thm:a-cert}
Let $F(x)$ be the true loss change of an implemented step $x$ and $\hat q(x)$ its predicted change, with $|F(x)-\hat q(x)|\le E(x)$ for a candidate $x_1$ and every comparator $x$ in a set $S$. If $\hat q(x_1)+E(x_1)<\hat q(x)-E(x)$ for every $x\in S$, then $F(x_1)<F(x)$ for every $x\in S$.
\end{atheorem}
The certificate needs no inverse of $H$. Each part of the error $E$ is bounded by a proved result:

\begin{atheorem}[The three error terms]\label{thm:a-terms}
(i)~\emph{Taylor remainder.} Let $\phi(t)$ be the loss after a fraction $t\in[0,1]$ of a step, twice continuously differentiable with $|\phi'''|\le M$ on $(0,1)$. Then $|\phi(1)-\phi(0)-\phi'(0)-\tfrac12\phi''(0)|\le M/6$.
(ii)~\emph{Estimation.} $|x^\top y|\le\|x\|\,\|y\|$. Hence, if $\|\hat g-g\|\le\delta_g$ and $|x^\top(\hat H-H)x|\le\delta_H\|x\|^2$ for all $x$, then $|\hat q(c)-q(c)|\le\delta_g\|c\|+\tfrac12\delta_H\|c\|^2$.
(iii)~\emph{Implementation.} If $\|\nabla L\|\le G$ on a convex set containing the endpoints of the exact and the implemented step, and both steps are norm-matched to radius $\rho$ from raw outputs $Y$ and $\hat Y$, then their losses differ by at most $2G\rho\,\|\hat Y-Y\|/\|Y\|$.
\end{atheorem}
In (i), $\phi'(0)=-g^\top c$ and $\phi''(0)=c^\top Hc$ are the first and second directional derivatives of the held-out loss along the step, which the diagnostic measures. Part (iii) combines the mean-value inequality with \cref{thm:a-transfer-err}.

\begin{acorollary}[End-to-end certificate]\label{cor:a-chain}
For an implemented candidate step and a comparator, let $I$, $T$ and $E$ bound the implementation, Taylor and estimation errors of each. If $\hat q_1+(I_1+T_1+E_1)<\hat q_0-(I_0+T_0+E_0)$, then the implemented candidate truly lowers the loss more than the comparator.
\end{acorollary}
\paragraph{Which inputs are established.} The theorems are unconditional; their \emph{inputs} differ in status. Established for our measurements: the trust-region conditions of \cref{thm:a-trust} ($\lambda>0$, residual below $10^{-15}$), so the reported best same-norm steps are certified optima of the model; the tangential gradient $p\neq0$ of \cref{thm:a-improve} ($\|p\|/\|r\|=\tanga$, \tangb, \tangc), so a same-norm improvement of the model exists at every checkpoint; and the hypotheses of \cref{thm:a-svd,thm:a-transfer-err}, which hold by construction. \emph{Not} established: the constants $M$, $\delta_g$, $\delta_H$ and $G$ of \cref{thm:a-terms}, and sampling intervals for the measured one-step losses. \Cref{cor:a-chain} is therefore not instantiated as a guarantee. The comparison in F2 with observed prediction errors is a consistency check only. \Cref{app:repro} specifies the measurements that would instantiate it.

\paragraph{Limits of the diagnostic.} Two further facts bound what the measurements can say.

\begin{aproposition}[Per-band ratios are not the fixed-norm criterion]\label{prop:a-ratio}
Let $H=\operatorname{diag}(1,3)$, $g=(\tfrac12,2)$, $u=(1,1)$ and $N=I$. The isolated ratio $g_b/H_{bb}$ is larger for band 2 ($\tfrac23$ against $\tfrac12$). Yet every same-norm step that gives band 2 more weight is strictly worse than $u$: $\|c\|=\|u\|$ and $c_2>1$ imply $q(c)>q(u)$. The same-norm step $c=(\tfrac{31}{25},\tfrac{17}{25})$, which gives band 1 more weight, is strictly better.
\end{aproposition}
On the circle, $q(c)-q(u)=\tfrac12(1-c_1)+(c_2-1)^2$. No cross-curvature is involved: at fixed norm what matters is the tangential part of $Hu-g$ (\cref{thm:a-improve}), not the ratio $g_b/H_{bb}$.

\begin{aproposition}[A one-step shape preference can reverse over two steps]\label{prop:a-horizon}
Let $F(x)=\tfrac12x_1^2+\tfrac14x_2^2$, whose gradient step of size 1 maps $x$ to $(0,\tfrac12x_2)$, and let $V(x)=F(0,\tfrac12x_2)$ be the loss after one more such step. The points $A=(\tfrac{142}{61},\tfrac{98}{61})$ and $B=(\tfrac{98}{61},\tfrac{142}{61})$ have the same norm as $x=(2,2)$. Then $F(B)<F(x)<F(A)$ but $V(A)<V(x)<V(B)$.
\end{aproposition}
Removing the step-size term therefore does not protect a one-step shape preference from horizon bias. This is why we treat the measured shape as a hypothesis for training experiments rather than as a prescription.

\subsection{Input-whitened Muon}\label{app:lean-white}

For a whitening factor $P\in\mathbb R^{n\times n}$, the whitened direction is $\mathcal W(G;P)=\mathrm{NS}(GP)\,P^\top$. With $PP^\top=C^{-1}$ it is the steepest direction for the data-dependent norm $\|DC^{1/2}\|_\op$.

\begin{atheorem}[Only the metric matters]\label{thm:a-factor}
If $P_1P_1^\top=P_2P_2^\top$ and $P_2$ is invertible, then $\mathcal W(G;P_1)=\mathcal W(G;P_2)$. The whitened step depends on $C$ only, not on the square root chosen.
\end{atheorem}
\emph{Proof idea.} $P_1=P_2Q$ with $Q$ orthogonal, and \cref{thm:a-ns} moves $Q$ through $\mathrm{NS}$.

\begin{atheorem}[Exact $\mathrm{GL}(n)$-equivariance]\label{thm:a-gl}
Reparametrize the inputs by an invertible $M$: $x'=Mx$, $W'=WM^{-1}$, $G'=GM^\top$, $C'=MCM^\top$.
(i)~If $PP^\top=C^{-1}$ and $P'$ is invertible with $P'P'^\top=C'^{-1}$, then $\mathcal W(G';P')=\mathcal W(G;P)\,M^{-1}$.
(ii)~The full step with decoupled weight decay is equivariant too:
\[
(1-\eta\lambda)WM^{-1}-\eta\,\mathcal W\big(GM^\top;(M^\top)^{-1}P\big)=\big[(1-\eta\lambda)W-\eta\,\mathcal W(G;P)\big]M^{-1}.
\]
The trained function is therefore independent of the input coordinates.
\end{atheorem}

\begin{atheorem}[Steepest directions transfer through a change of metric]\label{thm:a-transfer}
Let $P$ be invertible and $\mathcal N$ any norm-like function. If $D$ is a steepest direction for $\mathcal N$ at $GP$, then $DP^\top$ is a steepest direction at $G$ for the pulled-back function $\Delta\mapsto\mathcal N\big(\Delta(P^\top)^{-1}\big)$.
\end{atheorem}

\begin{atheorem}[Whitened Muon is conditioning-free on least squares]\label{thm:a-ls}
Let $L(W)=\tfrac12\operatorname{tr}\big((W-W_\star)C(W-W_\star)^\top\big)$ with $C=RR^\top$ and $R$ invertible. Run whitened Muon with $P=(R^\top)^{-1}$ and any step sizes $\eta_0,\eta_1,\dots$. Then the whitened error $E_t=(W_t-W_\star)R$ satisfies
\[
E_{t+1}=E_t-\eta_t\,\mathrm{NS}(E_t),\qquad L(W_t)=\tfrac12\|E_t\|_\F^2,
\]
which is plain Muon on the isotropic loss $\tfrac12\|E\|_\F^2$, for every $C$, however ill-conditioned.
\end{atheorem}

\begin{aproposition}[Plain Muon is not $\mathrm{GL}(n)$-equivariant]\label{prop:a-plain}
For $G=(1,1)$ and $M=\operatorname{diag}(2,1)$: if plain Muon's step were the same in both coordinate systems, $\mathrm{NS}(GM^\top)M=\mathrm{NS}(G)$, then $\mathrm{NS}(G)=0$.
\end{aproposition}

\subsection{Energy-matched decay}\label{app:lean-decay}

Ordinary decoupled decay $W\leftarrow W-\eta\lambda W$ shrinks every singular value by the same factor. Energy-matched Schatten-1 decay replaces $\lambda W$ by $\lambda\,(\|W\|_\F^2/\langle W,P\rangle)\,P$ with $P$ the polar factor of $W$. Since $\langle W,P\rangle=\|W\|_*=\sum_i\sigma_i$, every singular value then loses the same absolute amount $c_1=\lambda\sum_j\sigma_j^2/\sum_j\sigma_j$.

\begin{atheorem}[Energy match]\label{thm:a-energy-decay}
(i)~If $\sum_j\sigma_j\neq0$, then $\sum_i\sigma_i\,c_1=\sum_i\sigma_i\,(\lambda\sigma_i)$: subtracting $c_1$ from every singular value removes the same first-order energy as ordinary decay.
(ii)~More generally, for matrices $W$ and $D$ with $\langle W,D\rangle\neq0$, the rescaled direction $\tilde D=(\|W\|_\F^2/\langle W,D\rangle)\,D$ satisfies $\langle W,\tilde D\rangle=\langle W,W\rangle$ and $\langle W,\tilde D-W\rangle=0$. The energy-matched step is ordinary decay plus a step orthogonal to $W$.
\end{atheorem}

\begin{aproposition}[Which singular values shrink more]\label{prop:a-decay-shrink}
Let $\lambda>0$ and $\sigma_i>0$. Under energy-matched Schatten-1 decay, $\sigma_i$ shrinks by a larger fraction than under ordinary decay ($c_1/\sigma_i>\lambda$) if and only if $\sigma_i<\sum_j\sigma_j^2/\sum_j\sigma_j$. If $c_1>0$ and $0<\sigma_i<\sigma_j$, then $c_1/\sigma_j<c_1/\sigma_i$: smaller singular values always lose a larger fraction.
\end{aproposition}
The decay therefore prunes the bulk of the spectrum and spares its top. That is the opposite of Schatten-4 decay, which hurt in our experiments while Schatten-1 helped.

\subsection{Results from earlier formal work cited in \cref{sec:lean}}

These results come from the earlier, temporal-filter part of the project. Most are explicit counterexamples or elementary identities, and we state them for what they rule out.

\begin{aproposition}[Moments do not determine the momentum preference under $\NS$]\label{prop:a-moments}
(i)~Any risk that is affine in the first two moments of a finite noise law takes the same value on two laws with equal mass, mean and second moment.
(ii)~There are two explicit laws for the amplitude of rank-two gradient noise with the following properties: equal means, equal second moments, identical temporal covariance of the noise streams, and identical linearized risk curves $\beta\mapsto R_{\rm lin}(\beta)$ for every momentum $\beta$. Yet under the actual five-step $\NS$ map with stabilized normalization, the first law strictly prefers $\beta=0.95$ to $\beta=0.85$ and the second strictly prefers $0.85$ to $0.95$.
(iii)~Hence no rule that observes only the linearized risk curve can choose the better momentum for both laws.
\end{aproposition}

\begin{aproposition}[Spatial orientation decides the momentum preference]\label{prop:a-orient}
In an explicit full-rank construction with two consecutive updates, the signal path, drift, noise energy and time dependence are all matched. The Euclidean block risk is identical for every $\beta$ whether the noise is tangential or radial. After Newton--Schulz with any finite depth (including five), tangential noise makes $\beta=0.95$ strictly better than $\beta=0.85$, and radial noise makes $\beta=0.85$ strictly better than $\beta=0.95$.
\end{aproposition}

\begin{aproposition}[First-order alignment can mis-rank finite steps]\label{prop:a-chord}
Let $L(x,y)=x+2y+\tfrac12(x^2+10y^2)$. The unit steps $A=(-1,0)$ and $B=(0,-1)$ have equal length. At the origin, first-order analysis prefers $B$ (directional decrease 2 against 1), yet $L(A)-L(B)=-\tfrac72$, so $A$ is the better finite step.
\end{aproposition}

\begin{aproposition}[Curvature decides between the polar and gradient directions]\label{prop:a-blend}
Let a step $d$ change a diagonal quadratic by $\Delta(d)=-\langle g,d\rangle+\tfrac12\sum_ih_id_i^2$, with gradient $g=(7,1)$. The polar direction $d_P=(1,1)$ and the norm-matched gradient direction $d_G=(7/5,1/5)$ both have $\|d\|^2=2$.
(i)~With curvature $h=(1,1)$, $d_G$ is better.
(ii)~With $h=(6,1)$, $d_P$ is better.
(iii)~With $h=(4,1)$, the norm-matched intermediate direction $(17/13,\,7/13)$ is strictly better than both.
\end{aproposition}
This is the two-direction version of \cref{fig:anatomy}a: which spectral allocation is best depends on how curvature compares with gradient signal across directions.

\begin{aproposition}[Greedy step-size facts]\label{prop:a-greedy}
For a quadratic loss with gradient $W\mapsto HW-b$ and a step $W-\eta O$:
(i)~$\langle\nabla L(W-\eta O),O\rangle=\langle\nabla L(W),O\rangle-\eta\,\langle HO,O\rangle$, which is affine in $\eta$.
(ii)~With $a=\langle\nabla L(W),O\rangle$ and $q=\langle HO,O\rangle>0$, the one-step loss $L(W-\eta O)$ decreases with $\eta$ exactly when $\eta<a/q$.
(iii)~With no aligned signal ($a=0$), every positive step looks too large to a greedy criterion.
(iv)~A bias added equally to the marginal value of every group does not change their comparison.
\end{aproposition}

\section{Diagnostic details and additional results}\label{app:diag}

\paragraph{Scope.} The diagnostic perturbs only the 48 hidden matrices, along exact-polar band directions or the actual $\NS$ neuron groups. The AdamW update of embeddings and norms and the weight decay are left out. $g$ and $H$ are estimated on separate held-out folds, so they are estimates of a common local model with separate errors, not derivatives of one empirical loss. The realized-loss checks use a third fold. The inverse-based quantities ($c^\star$ and the shape shares) are sensitive because $H$ is ill-conditioned. The fixed-norm band-filter predictions evaluate $q$ at a given vector and need no inverse.

\paragraph{Procedure.} For a checkpoint at step $t$ (Muon, seed 43, screen schedule), we recompute the next training minibatch's gradient with the same clipping, form $N=G+\beta(\beta B+G)$ exactly as the optimizer would, and take the full SVD of $N$ for each of the 48 matrices. Band directions use the exact polar factor $U_{I_b}V_{I_b}^\top$, scaled by the learning rate at step $t$ and Muon's shape factor. Neuron groups use Muon's actual $\NS$ step, split into six equal-count sextiles by row norm (tall or square matrices) or column norm (wide matrices). The held-out gradient uses 768 validation sequences (393k tokens). Each Hessian--vector product uses 192 further sequences (98k tokens), computed by double backward with the math attention kernel. Realized loss changes use 768 further sequences. The three sets are disjoint. For $B=6$ groups and five group sets (all matrices, and each layer type), the diagnostic needs 30 Hessian--vector products per checkpoint.

\begin{table}[h]
\centering
\caption{Band diagnostic over all 48 matrices. Top: per-band multiplier $c_b=g_b/H_{bb}$, best uniform rescaling $\gamma^\star$, second-order gap $q(\mathbf 1)-q(c^\star)$ (nats) and its shape share (\cref{thm:split}). Bottom: the joint optimum $c^\star=H^{-1}g$ and the condition number of $H$. $c^\star$ is irregular because $H$ is ill-conditioned, but it also puts nearly all its weight on the lower half of the spectrum. All multipliers scale as $1/\eta$; $\eta$ at step 2,048 is $10\times$ smaller.}
\label{tab:diag}
\small
\begin{tabular}{@{}lcccccc|ccc@{}}
\toprule
Step & 1--8 & 9--32 & 33--96 & 97--256 & 257--512 & 513--768 & $\gamma^\star$ & gap & shape share \\
\midrule
1,024 & 0.59 & 1.96 & 2.46 & 4.27 & 12.0 & 28.3 & 0.52 & 0.00674 & 17\% \\
1,536 & 0.55 & 1.90 & 2.39 & 4.21 & 9.52 & 21.0 & 0.49 & 0.00532 & 12\% \\
2,048 & 0.98 & 7.67 & 16.4 & 31.6 & 75.8 & 186.3 & 1.52 & 0.00137 & 97\% \\
\midrule
1,024 ($c^\star$) & 0.53 & 0.18 & $-$0.10 & 0.31 & 5.92 & 16.4 & \multicolumn{3}{l}{cond$(H)\approx1\times10^{4}$} \\
1,536 ($c^\star$) & 0.48 & 0.34 & $-$0.15 & 0.88 & 4.01 & 11.5 & \multicolumn{3}{l}{cond$(H)\approx8\times10^{3}$} \\
2,048 ($c^\star$) & 0.40 & 0.12 & 3.16 & 8.05 & 34.8 & 115.9 & \multicolumn{3}{l}{cond$(H)\approx4\times10^{3}$} \\
\bottomrule
\end{tabular}

\end{table}

\begin{table}[h]
\centering
\caption{Quadratic-model calibration (bands, all matrices): predicted versus realized change in held-out loss (393k disjoint tokens) for Muon's step ($c=\mathbf 1$) and the model-optimal band shape ($c=c^\star$).}
\label{tab:calib}
\small
\begin{tabular}{@{}lcccc@{}}
\toprule
Checkpoint & Muon, predicted & Muon, realized & $c^\star$, predicted & $c^\star$, realized\\
\midrule
1,024 & \preMuona & \actMuona & \preStara & \actStara\\
1,536 & \preMuonb & \actMuonb & \preStarb & \actStarb\\
2,048 & \preMuonc & \actMuonc & \preStarc & \actStarc\\
\bottomrule
\end{tabular}
\end{table}

\paragraph{Measured one-step losses at matched norm.} To test the fixed-norm predictions with the actual loss rather than the quadratic model, we start from the same checkpoints and form each matrix's next Nesterov input. We take the exact tilted direction $\phi_5(\sigma)(\sigma/\sigma_{\rm rms})^p$, rescaled per matrix to the Frobenius norm of Muon's $\NS$ step ($p=0$ is Muon), or a norm-matched band filter. We then measure the held-out loss after one step on a fixed 393k-token set (\cref{tab:tiltslope}). The slope $dL/dp$ at $p=0$ is a central finite difference. The best same-norm filter is the step-1,536 model optimum. One checkpoint seed and one validation subset are used, without intervals.

\begin{table}[h]
\centering
\caption{Measured change of the held-out loss ($\times10^{-3}$) after one step at Muon's own update norm: the exact tilt $\sigma^p$ for several $p$, halving ranks 1--8, and the best same-norm filter, plus the slope $dL/dp$ at $p=0$.}
\label{tab:tiltslope}
\small
\begin{tabular}{@{}lrrrrrrrrr@{}}
\toprule
Step & $p=-\frac12$ & $-\frac14$ & $-0.1$ & $0$ (Muon) & $+0.1$ & $+\frac14$ & ranks 1--8 $\times\frac12$ & best same-norm & $dL/dp|_0$ \\
\midrule
1,024 & $-$3.07 & $-$5.75 & $-$5.93 & $-$1.76 & $+$12.08 & $+$80.45 & $-$6.14 & $-$6.28 & $+$0.090 \\
1,536 & $-$2.48 & $-$4.13 & $-$3.99 & $-$1.03 & $+$7.95 & $+$48.44 & $-$4.27 & $-$4.40 & $+$0.060 \\
2,048 & $-$0.13 & $-$0.26 & $-$0.32 & $-$0.33 & $-$0.38 & $-$0.22 & $-$0.29 & $-$0.21 & $-$0.000 \\
\bottomrule
\end{tabular}

\end{table}

\paragraph{Fixed-norm analysis.} Band directions $D_{\ell,b}=\eta s_\ell U_{I_b}V_{I_b}^\top$ are mutually Frobenius-orthogonal, with $\|D_b\|_\F^2\propto|I_b|$ (every hidden matrix has rank 768). Every matrix's step therefore has the same norm as Muon's exactly when $c^\top Nc=\mathbf 1^\top N\mathbf 1$ with $N=\operatorname{diag}(8,24,64,160,256,256)$. The best same-norm step minimizes $q$ on this ellipsoid. With $\tilde H=N^{-1/2}HN^{-1/2}$ and $\tilde g=N^{-1/2}g$, it is $c=N^{-1/2}(\tilde H+\lambda I)^{-1}\tilde g$, with $\lambda>-\lambda_{\min}(\tilde H)$ chosen by a one-dimensional root search so that the constraint holds. Norm-matched filters are rescaled onto the ellipsoid before evaluation. \Cref{tab:fixed} gives the predictions.

\begin{table}[h]
\centering
\caption{Second-order predictions at Muon's own update norm (bands, all 48 matrices). Column 2 is Muon's predicted one-step change $q(\mathbf 1)$. Columns 3--5 are $q(c)-q(\mathbf 1)$ for the best same-norm step, the norm-matched band filter used in training ($m_b\propto c_b^{1/4}$ at step 1,024), and halving ranks 1--8 with norm matching. The last column is the relative size of the tangential gradient $p$ (\cref{thm:a-improve}). Bottom: the best same-norm step's band multipliers.}
\label{tab:fixed}
\small
\begin{tabular}{@{}lccccc@{}}
\toprule
Checkpoint & Muon's $q(\mathbf 1)$ & best same-norm $c$ & band filter & ranks 1--8 halved & $\|p\|/\|r\|$ \\
\midrule
1,024 & $-$0.00079 & $-$0.00589 & $-$0.00570 & $-$0.00527 & 1.00 \\
1,536 & $+$0.00026 & $-$0.00488 & $-$0.00478 & $-$0.00443 & 1.00 \\
2,048 & $-$0.00028 & $-$0.00009 & $+$0.00005 & $+$0.00001 & 0.79 \\
\midrule
\multicolumn{6}{@{}l}{Best same-norm $c$ at step 1,024: (0.54, 0.22, 0.21, 1.26, 1.26, 0.62)} \\
\multicolumn{6}{@{}l}{Best same-norm $c$ at step 1,536: (0.49, 0.35, 0.17, 1.38, 1.17, 0.63)} \\
\multicolumn{6}{@{}l}{Best same-norm $c$ at step 2,048: (0.64, 2.92, 2.30, 1.05, 0.39, 0.17)} \\
\bottomrule
\end{tabular}

\end{table}

\begin{figure}[h]
\centering
\begin{minipage}[t]{0.47\textwidth}\centering
\includegraphics[width=\textwidth]{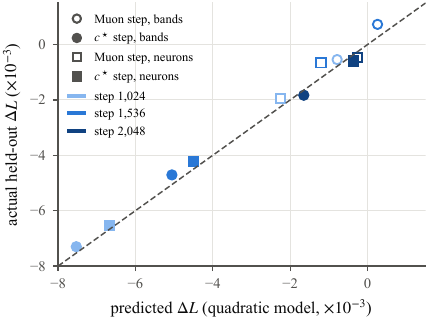}
\caption{Predicted (quadratic model) versus realized held-out loss change for 12 test steps.}
\label{fig:calibration}
\end{minipage}\hfill
\begin{minipage}[t]{0.47\textwidth}\centering
\includegraphics[width=\textwidth]{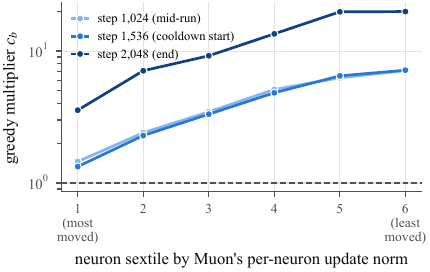}
\caption{Greedy multiplier per neuron sextile, ranked by Muon's per-neuron update norm. The most-moved neurons sit in higher-curvature directions.}
\label{fig:neurons}
\end{minipage}
\end{figure}

\begin{figure}[h]
\centering
\includegraphics[width=\textwidth]{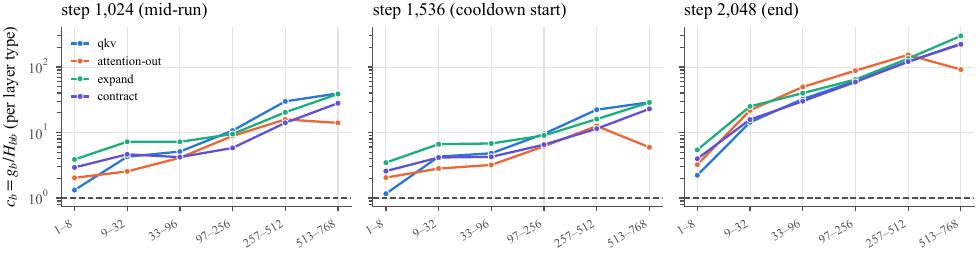}
\caption{Band multipliers computed separately for each layer type. The top band has the smallest multiplier in every case.}
\label{fig:bandtypes}
\end{figure}

\paragraph{Implied spectral exponent.} To relate the band multipliers to a power law $f(\sigma)\propto\sigma^p$, we take each matrix's singular values at ranks 1, 8, 32, 96 and 256, normalize by the value at rank 96, and average them in log space over the 48 matrices. Each of the top four bands is represented by the geometric mean of its two edges, and we regress $\log c_b$ on $\log\sigma_b$. The slopes are \impA, \impB{} and \impC{} at the three checkpoints. The estimate is crude: the band representatives are coarse, and the tail bands, whose lower edges we did not log, are excluded. It is also a greedy quantity, so it overstates how strong a tilt to use. We report it only to show that the prescribed sign is negative and of order one in magnitude. The deployed power, $-\tfrac14$, is deliberately conservative.

\paragraph{Per-neuron spread.} \Cref{tab:leverage} reports the coefficient of variation (CV) of per-neuron update norms at the three checkpoints. Each entry is the mean over the 12 matrices of a type, along the axis that indexes neurons: rows for qkv, expand and attention-out, columns for contract. For comparison, the raw momentum's QKV row-norm CV is \levMomQKV{}. Muon reduces the spread but does not remove it. Block-wise $\NS$ steps leave a residual QKV spread (CV \levQKVsq) although exact orthogonal blocks would have none (\cref{prop:a-rows}). A plausible cause is that $\NS$ maps the near-null singular directions of low-rank blocks to values well below 1; we have not verified this.
\begin{table}[h]
\centering
\caption{Coefficient of variation of per-neuron update norms of Muon's step and of the block-wise polar step.}
\label{tab:leverage}
\small
\begin{tabular}{@{}lcccc@{}}
\toprule
Layer (neuron axis) & Muon, step 1,024 & Muon, step 2,048 & Block-wise, step 1,024 & Muon max/median, step 1,024\\
\midrule
qkv (rows) & \levQKV & \levQKVend & \levQKVsq & \levQKVmax\\
expand (rows) & \levExp & 0.18 & \levExpSq & 1.72\\
contract (columns) & \levCon & 0.17 & \levConSq & 1.67\\
attention-out (rows) & \levAttn & 0.02 & \levAttn & 1.05\\
\bottomrule
\end{tabular}
\end{table}

\section{All screen runs}\label{app:screen}

\begin{figure}[h]
\centering
\includegraphics[width=0.95\textwidth]{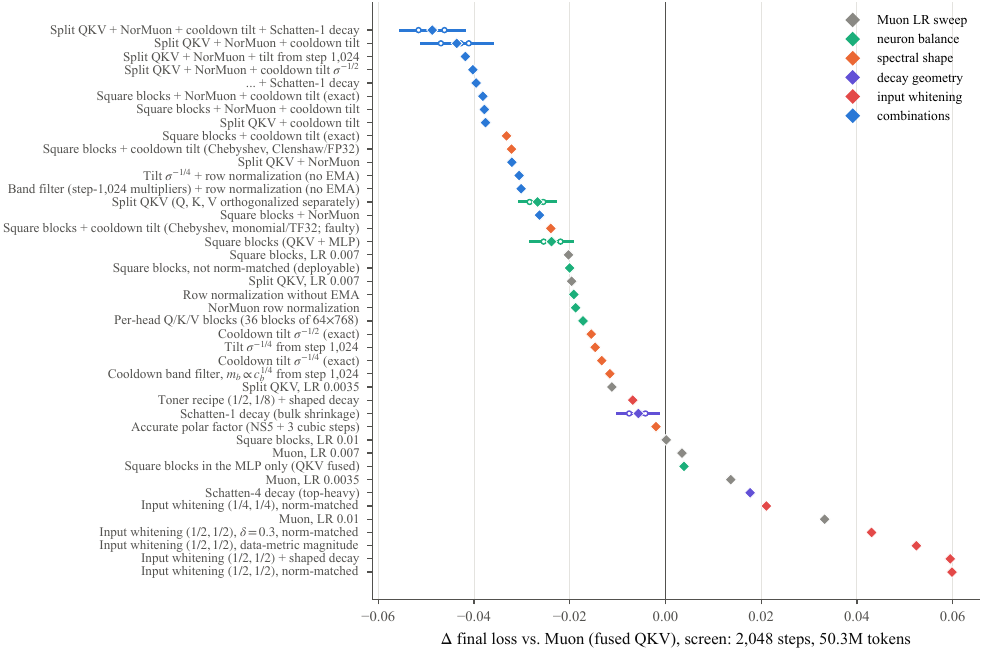}
\caption{Screen arms sorted by their mean paired difference from fused-QKV Muon (seed 42, or the three-seed mean with 95\% $t$-interval and open circles for seeds where three seeds exist).}
\label{fig:screen}
\end{figure}

\Cref{tab:screen,fig:screen} list every completed screen arm (2,048 steps, 50.3M tokens). Differences are paired against fused-QKV Muon at LR 0.005: the seed-42 difference for every arm, and the mean over seeds 42--44 with a 95\% $t$-interval where three seeds exist. The arms were selected adaptively during the study; no multiplicity correction is applied.

\begin{table}[h]
\centering
\caption{Screen results. Final held-out loss (4.18M tokens) at seed 42 and paired differences from fused-QKV Muon.}
\label{tab:screen}
\scriptsize
\begin{tabular}{@{}llccc@{}}
\toprule
Family & Arm & Final (s42) & $\Delta$ (s42) & $\Delta$, 3 seeds [95\% CI] \\
\midrule
 & Muon, LR 0.005 (seeds 42/43/44: 4.2043, 4.2066, 4.2035) & 4.2043 & --- & --- \\
\midrule
Learning rate & Muon, LR 0.0035 & 4.2179 & $+$0.0136 &  \\
 & Muon, LR 0.007 & 4.2077 & $+$0.0034 &  \\
 & Muon, LR 0.01 & 4.2376 & $+$0.0332 &  \\
 & Split QKV, LR 0.0035 & 4.1931 & $-$0.0112 &  \\
 & Split QKV, LR 0.007 & 4.1847 & $-$0.0196 &  \\
 & Square blocks, LR 0.007 & 4.1840 & $-$0.0203 &  \\
 & Square blocks, LR 0.01 & 4.2045 & $+$0.0001 &  \\
\midrule
Neuron balance & Split QKV (Q, K, V orthogonalized separately) & 4.1788 & $-$0.0255 & $-$0.0267 [$-$0.0304, $-$0.0230] \\
 & Per-head Q/K/V blocks (36 blocks of $64{\times}768$) & 4.1871 & $-$0.0172 &  \\
 & Square blocks (QKV + MLP) & 4.1824 & $-$0.0219 & $-$0.0238 [$-$0.0282, $-$0.0194] \\
 & Square blocks, not norm-matched (deployable) & 4.1843 & $-$0.0200 &  \\
 & Square blocks in the MLP only (QKV fused) & 4.2082 & $+$0.0039 &  \\
 & NorMuon row normalization & 4.1854 & $-$0.0189 & $-$0.0188 [$-$0.0193, $-$0.0183] \\
 & Row normalization without EMA & 4.1852 & $-$0.0192 &  \\
\midrule
Spectral shape & Cooldown tilt $\sigma^{-1/4}$ (exact) & 4.1910 & $-$0.0133 &  \\
 & Cooldown tilt $\sigma^{-1/2}$ (exact) & 4.1888 & $-$0.0155 &  \\
 & Tilt $\sigma^{-1/4}$ from step 1,024 & 4.1896 & $-$0.0147 &  \\
 & Cooldown band filter, $m_b\propto c_b^{1/4}$ from step 1,024 & 4.1927 & $-$0.0116 &  \\
 & Accurate polar factor (NS5 + 3 cubic steps) & 4.2023 & $-$0.0020 &  \\
 & Square blocks + cooldown tilt (Chebyshev, monomial/TF32; faulty) & 4.1804 & $-$0.0240 &  \\
 & Square blocks + cooldown tilt (Chebyshev, Clenshaw/FP32) & 4.1721 & $-$0.0322 &  \\
 & Square blocks + cooldown tilt (exact) & 4.1711 & $-$0.0332 &  \\
\midrule
Decay geometry & Schatten-1 decay (bulk shrinkage) & 4.1991 & $-$0.0052 & $-$0.0057 [$-$0.0100, $-$0.0014] \\
 & Schatten-4 decay (top-heavy) & 4.2220 & $+$0.0177 &  \\
\midrule
Data metric (whitening) & Input whitening $(\frac14,\frac14)$, norm-matched & 4.2254 & $+$0.0211 &  \\
 & Input whitening $(\frac12,\frac12)$, norm-matched & 4.2642 & $+$0.0599 &  \\
 & Input whitening $(\frac12,\frac12)$, $\delta{=}0.3$, norm-matched & 4.2474 & $+$0.0431 &  \\
 & Input whitening $(\frac12,\frac12)$, data-metric magnitude & 4.2567 & $+$0.0524 &  \\
 & Input whitening $(\frac12,\frac12)$ + shaped decay & 4.2638 & $+$0.0595 &  \\
 & Toner recipe $(\frac12,\frac18)$ + shaped decay & 4.1975 & $-$0.0068 &  \\
\midrule
Combinations & Tilt $\sigma^{-1/4}$ + row normalization (no EMA) & 4.1737 & $-$0.0306 &  \\
 & Band filter (step-1,024 multipliers) + row normalization (no EMA) & 4.1742 & $-$0.0302 &  \\
 & Square blocks + NorMuon & 4.1780 & $-$0.0263 &  \\
 & Square blocks + NorMuon + cooldown tilt (exact) & 4.1661 & $-$0.0382 &  \\
 & Square blocks + NorMuon + cooldown tilt & 4.1665 & $-$0.0378 &  \\
 & ... + Schatten-1 decay & 4.1648 & $-$0.0396 &  \\
 & Split QKV + cooldown tilt & 4.1667 & $-$0.0376 &  \\
 & Split QKV + NorMuon & 4.1722 & $-$0.0321 &  \\
 & Split QKV + NorMuon + cooldown tilt & 4.1632 & $-$0.0411 & $-$0.0436 [$-$0.0511, $-$0.0362] \\
 & Split QKV + NorMuon + tilt from step 1,024 & 4.1625 & $-$0.0418 &  \\
 & Split QKV + NorMuon + cooldown tilt $\sigma^{-1/2}$ & 4.1641 & $-$0.0402 &  \\
 & Split QKV + NorMuon + cooldown tilt + Schatten-1 decay & 4.1581 & $-$0.0462 & $-$0.0487 [$-$0.0555, $-$0.0420] \\
\bottomrule
\end{tabular}

\end{table}

\begin{table}[h]
\centering
\caption{\textbf{Theory versus training.} What the Lean statements certify, whether the anatomy implies a gain, and the measured paired $\Delta$ against fused-QKV Muon (screen, one seed, unless marked). Formal properties alone did not predict which changes help.}
\label{tab:theory}
\small
\setlength{\tabcolsep}{3pt}
\begin{tabularx}{0.75\textwidth}{@{}>{\raggedright\arraybackslash}p{1.9cm}>{\raggedright\arraybackslash}X>{\centering\arraybackslash}p{0.7cm}>{\raggedleft\arraybackslash}p{1.55cm}@{}}
\toprule
Change & Certified property & Anat.\ & $\Delta$ \\
\midrule
Split QKV & steepest for block max-norm; exact-polar blocks have unit rows; energy identity & yes & \dSvM$^{\dagger}$\\
MLP square blocks & same & weak & \scrMLPBlocks\\
Cooldown tilt & keeps Muon's $O(n)$-equivariance & yes & \dSTvS$^{\dagger\ddagger}$\\
Accurate polar & --- & no & \scrPolish\\
Input whitening & $\mathrm{GL}(n)$-equivariance; conditioning-free LS (undamped form) & --- & $+$0.021--0.060\\
Toner recipe & (prior art) & --- & \scrToner\\
Schatten-4 decay & --- & --- & \scrSchFour\\
Schatten-1 decay & --- & --- & \scrSchOneMean$^{\S}$\\
\bottomrule
\end{tabularx}
\\[1pt]{\footnotesize $^\dagger$100.7M tokens, $n=3$.\ \ $^\ddagger$vs.\ split-QKV Muon.\ \ $^\S n=3$.}
\end{table}

\paragraph{Learning-rate robustness of block-wise steps.} With square blocks, raising the learning rate from 0.005 to 0.007 and 0.01 gives final losses of 4.1840 and 4.2045. Muon at the same rates gives 4.2077 and 4.2376 (\scrLRc{} and \scrLRd{} relative to its optimum). Doubling the learning rate leaves square-block Muon level with Muon's tuned optimum. MuonBP reported that block-only training needed smaller learning rates at scale~\citep{muonbp}; in our setting block-wise steps are, if anything, more tolerant of large rates.

\section{Implementation details}\label{app:impl}

\begin{algorithm}[h]
\caption{One step for a hidden matrix: split-QKV Muon with NorMuon row normalization and the cooldown tilt (\texttt{cooldown} is true in the last 25\% of steps).}
\label{alg:step}
\begin{algorithmic}[1]
\small
\STATE $B\leftarrow\beta B+G$;\ \ $N\leftarrow G+\beta B$
\STATE Split $N$ into blocks $N_j$ (Q, K, V for the attention input matrix; one block otherwise)
\FOR{each block $N_j$}
  \STATE $O_j\leftarrow\NS(N_j)$
  \IF{\texttt{cooldown}}
    \STATE $S\leftarrow \mathrm{sym}(O_jN_j^\top)$;\ \ $\lambda\leftarrow 1.01\,\lambda_{\max}(S)$ (20 power iters)
    \STATE $D_j\leftarrow\mathrm{ClenshawCheb}_{x^{-1/4}}(S/\lambda)\,O_j$ \hfill{\footnotesize (FP32)}
  \ELSE
    \STATE $D_j\leftarrow O_j$
  \ENDIF
\ENDFOR
\STATE $D\leftarrow[D_1;\dots]$;\ \ $v\leftarrow\beta_2v+(1-\beta_2)\,\mathrm{rowmean}(D\odot D)$;\ \ $D\leftarrow D/(\sqrt v+\epsilon)$
\STATE $D\leftarrow D\cdot\|\NS(N)\|_\F/\|D\|_\F$ \hfill{\footnotesize (norm match)}
\STATE $W\leftarrow(1-\eta\lambda)W-\eta\, s\, D$
\end{algorithmic}
\end{algorithm}

\paragraph{Norm matching.} For every matrix and step, the candidate direction $D$ is rescaled to $\|\NS(N)\|_\F$, with $N$ the same momentum Muon would use, and then enters \eqref{eq:muon} with Muon's shape factor and learning rate. Norm-matched whitening arms are matched to $\|\NS(NP)\|_\F$, the norm of their own whitened polar factor. The trainer enables TF32 matrix products globally, so $\NS$ runs in FP32 storage with TF32 products; only the Chebyshev filter disables TF32. Matching costs one extra joint $\NS$ per matrix for block-wise arms, used only to read its norm. The deployable square-block step omits it: it scales wide-matrix column blocks by $1/\sqrt k$ and runs all 144 square blocks in one batched $\NS$ call. It gives \scrBlocksDeploy{} on the screen, against \scrBlocks{} norm-matched.

\paragraph{NorMuon.} Following \citet{normuon}, the orthogonalized update is divided row-wise by $\sqrt{v}+10^{-8}$, where $v$ is an EMA ($\beta_2=0.95$) of the row mean of $D\odot D$. It is then norm-matched. A stateless variant with $\beta_2=0$ gives \scrRowInst{}, against \scrNor{} for the EMA version (screen), so the EMA is not what helps.

\paragraph{Spectral tilt.} The exact tilt maps singular value $\sigma_i$ of $N$ to $\phi_5(\sigma_i/\|N\|_\F)\cdot(\sigma_i/\sigma_{\rm rms})^p$ via an eigendecomposition of $NN^\top$. With $p=0$ it reproduces $\NS$ to $2\times10^{-4}$ relative error. The matmul-only tilt (\cref{alg:step}) uses a degree-12 Chebyshev least-squares fit of $x^p$ on $[10^{-3},1]$ (4,000 Chebyshev nodes) in the variable $t=2(x-x_{\rm lo})/(1-x_{\rm lo})-1$. It is evaluated by Clenshaw's recurrence with TF32 disabled. $S=ON^\top$ is symmetrized and scaled by $\Lambda=1.01$ times a 20-step power-iteration estimate of $\lambda_{\max}(S)$. That is not a proved upper bound: eigenvalues above $\Lambda$, or below $10^{-3}\Lambda$, fall outside the fitted interval, where the Chebyshev polynomial is not controlled. By \cref{thm:a-svd}, the implemented filter is $f_i^{1+p}\sigma_i^p$ rather than the exact $f_i\sigma_i^p$; on the Newton--Schulz plateau they differ by a factor in $[0.96,1.10]$. On real momenta the matmul tilt matches the exact tilt to direction cosine 0.998--0.999; enabling TF32 lowers the worst block to 0.992. By \cref{thm:a-transfer-err}(ii) the normalized steps differ by at most $6.3\%$ in Frobenius norm. Turning that into a loss bound would need a curvature bound along the difference (\cref{thm:a-cert}), which we do not have. When Q, K and V are split, the three $768\times768$ blocks of all 12 layers and the square attention-output matrices are filtered in one batched chain. The MLP matrices are filtered whole. With TF32 off the tilt costs about 0.17\,s per step on an A6000, during the cooldown only.

\paragraph{Input whitening.} $C$ is an EMA of the uncentered input second moment, trace-normalized with damping, $\hat C=C/\overline{\lambda(C)}+\delta I$ ($\delta=0.1$ unless stated). It is refreshed from 4,096-token activation samples. The step is $\polar(N\hat C^{-q_{\rm in}})\hat C^{-q_{\rm out}}$: $(\tfrac14,\tfrac14)$ and $(\tfrac12,\tfrac12)$ norm-matched, $(\tfrac12,\tfrac12)$ at its own magnitude, and the Toner recipe $(\tfrac12,\tfrac18)$ with shaped decay $\lambda W\hat C^{-1/4}$.

\paragraph{Schatten-$p$ decay.} We replace $\lambda W$ by $\lambda\frac{\|W\|_\F^2}{\langle W,\,W(W^\top W)^{(p-2)/2}\rangle}W(W^\top W)^{(p-2)/2}$. This removes the same Frobenius energy per step as uniform decay, with relative rate $\lambda\sigma_i^{p-2}\sum\sigma^2/\sum\sigma^p$ on singular value $\sigma_i$. For $p=1$ we use $\NS(W)$ for the polar factor.

\paragraph{Cost.} Measured seconds per step on one RTX A6000 (100.7M-token runs): Muon 0.343; NorMuon 0.329; norm-matched square blocks 0.415; square blocks $+$ NorMuon $+$ tilt 0.418 before and 0.588 during the cooldown. Mean wall time per 100.7M-token run, including evaluations: Muon \minM{} min, split-QKV \minS, split-QKV $+$ tilt \minST, split-QKV $+$ NorMuon \minSN, split-QKV $+$ NorMuon $+$ tilt \minSNT. Three runs shared a host, so these timings are indicative.

\section{Reproducibility and provenance}\label{app:repro}

\paragraph{Logging.} Each run writes an event log with its recipe, optimizer variant, seed, software versions and a hash of the optimizer source. The results tables, figures and run-derived numbers in the results text are generated automatically from these logs. A few descriptive quantities (per-step timings, curvature ratios, some leverage-table entries and filter cosines) were transcribed and checked in an independent pass against the logs. Training uses \texttt{torch.compile} and BF16 autocast for the forward pass; the optimizer state is FP32, and Newton--Schulz runs in FP32 with TF32 matrix products. The trainer reproduces an earlier, slower harness's Muon loss (3.9445 against 3.9463 at seed 42).

\paragraph{Order of experiments.} The screen arms, the diagnostic and the first confirmations ran in parallel on three GPUs over one day. The anatomy was \emph{not} pre-registered: the square-block, whitening and accurate-polar screens started before the curvature diagnostics, and the neuron diagnostic postdates the first block arms. The square-block arm was motivated by the leverage analysis; the attribution runs later showed that the QKV split carries its gain. Two confirmations, square blocks $+$ NorMuon $+$ tilt and split QKV $+$ NorMuon $+$ tilt, were launched before their own screens had finished. The 355M seed-42 pair was launched after the positive 124M confirmation. The gating rule of \cref{sec:experiments} was adopted after that; it released the full-length factorial ablations (split QKV $+$ tilt; split QKV $+$ NorMuon), whose screens beat the baseline by 0.0121 and 0.0066, and the 355M seeds 43--44.

\paragraph{Source versions.} The optimizer file changed during the study. The original fused-QKV Muon, NorMuon and square-block confirmations span four source hashes, and a read-only audit of the raw logs could not certify, from the retained files, that the changes touched only inactive branches. All split-QKV arms (split-QKV Muon, the factorial and the 355M runs) ran on a single source version, so the factorial comparisons are unaffected. Every comparison involving fused-QKV Muon, NorMuon on fused QKV, square blocks or square blocks $+$ NorMuon $+$ tilt crosses source versions; this includes the attribution results ``square blocks vs.\ split QKV'' and ``NorMuon vs.\ split QKV''. The A/A bound and the reproduction of the earlier harness's loss suggest the effect is small, but we cannot certify it.

\paragraph{Measurements needed to instantiate the certificate.} Experiments were paused before these could run. Each uses saved checkpoints and no training:
(1)~\emph{Sampling intervals}: record per-sequence held-out losses for every step in \cref{tab:tiltslope} and bootstrap paired differences over sequences, across several checkpoint seeds.
(2)~\emph{Estimation errors} $\delta_g,\delta_H$: recompute $g$ and $H$ on disjoint folds and take the spread of $\hat g$ and of $c^\top\hat Hc$ over folds.
(3)~\emph{Third-derivative bound} $M$: evaluate the loss at $t\in\{0,\pm\tfrac12,\pm1\}$ along each step and take third finite differences.
(4)~\emph{Gradient bound} $G$: the held-out gradient norm at both endpoints and the midpoint of each step pair.

\paragraph{Statistics.} Intervals are two-sided 95\% Student-$t$ intervals over three paired seeds (2 degrees of freedom). They are nominal: the same seeds were used in the screen, and no correction is made for the number of arms examined. The A/A bound (\aaMax\nats{}) uses 98k-token intermediate evaluations; final evaluations use 4.18M tokens and are less noisy.

\end{document}